\documentclass{article}
\usepackage{graphicx}

\usepackage{times}
\usepackage{helvet}
\usepackage{courier}

\newcommand{\eq}[1]{Eq.~(\ref{eq.#1})} 

\newcommand{\fig}[1]{Fig.~\ref{fig.#1}}

\newcommand{\tbl}[1]{Table~\ref{table.#1}}

\newcommand{\sect}[1]{Section~\ref{sect.#1}}

\newcommand{\sectlabel}[1]{\label{sect.#1}}
\newcommand{\eqlabel}[1]{\label{eq.#1}}
\newcommand{\figlabel}[1]{\label{fig.#1}}
\newcommand{\tbllabel}[1]{\label{table.#1}}

\newcommand{\figwidth}{4in}
\newcommand{\figwidthS}{2.5in}

\newcommand{\vAvg}{v_{\rm{avg}}} 
\newcommand{\vAvgCell}{v_{\rm{cell}}} 
\newcommand{\pressure}{p}
\newcommand{\power}{P}
\newcommand{\pressureGradient}{\nabla \pressure} 
\newcommand{\Doxygen}{D_{\mbox{\scriptsize \Oxygen}}} 
\newcommand{\Dheme}{D_{\mbox{\scriptsize heme}}}

\newcommand{\vFluid}{{\bf v}}
\newcommand{\Flux}{{\bf F}}
\newcommand{\totalOxygenFlux}{\ensuremath{J_{\mbox{\scriptsize \Oxygen}}}} 

\newcommand{\reactionSiteDensity}{\ensuremath{\rho_{\mbox{\scriptsize react}}}}    
\newcommand{\reactionSiteNumber}{\ensuremath{N_{\mbox{\scriptsize react}}}}    
\newcommand{\reactionRate}{\ensuremath{r}}   
\newcommand{\ChalfReaction}{\ensuremath{K}}  
\newcommand{\reactionEnergy}{\ensuremath{e}} 
\newcommand{\PowerRobot}{\ensuremath{\power_{\mbox{\scriptsize robot}}}}

\newcommand{\PowerTissue}{\ensuremath{\power_{\mbox{\scriptsize tissue}}}}
\newcommand{\PowerTissueMax}{\ensuremath{\power^{\mbox{\scriptsize max}}_{\mbox{\scriptsize tissue}}}}
\newcommand{\ChalfReactionTissue}{K_{\mbox{\scriptsize tissue}}}
\newcommand{\Rtissue}{R_{\mbox{\scriptsize tissue}}}
\newcommand{\Rcell}{R_{\mbox{\scriptsize cell}}} 
\newcommand{\Rrobot}{L_{\mbox{\scriptsize robot}}} 
\newcommand{\Vrobot}{V_{\mbox{\scriptsize robot}}} 

\newcommand{\BoltzmannConstant}{\ensuremath{k_B}}

\newcommand{\density}{\ensuremath{\rho}}
\newcommand{\viscosity}{\ensuremath{\eta}}
\newcommand{\kThermal}{\ensuremath{k_{\mbox{\scriptsize thermal}}}}
\newcommand{\heatCapacity}{\ensuremath{c_{\mbox{\scriptsize
thermal}}}}

\newcommand{\nHill}{n}  
\newcommand{\Phalf}{\pressure_{50}} 
\newcommand{\tUnloading}{t_u}
\newcommand{\Oxygen}{\ensuremath{\mbox{O}_2}}
\newcommand{\CarbonDioxide}{\ensuremath{\mbox{CO}_2}}
\newcommand{\CoxygenMax}{C_{\mbox{\scriptsize \Oxygen}}^{\rm max}}
\newcommand{\Coxygen}{C_{\mbox{\scriptsize \Oxygen}}}
\newcommand{\Hoxygen}{H_{\mbox{\scriptsize \Oxygen}}} 
\newcommand{\Sequib}{S_{\mbox{\scriptsize equib}}} 
\newcommand{\hematocrit}{\ensuremath{h}} 
\newcommand{\hematocritFull}{\ensuremath{h_{\mbox{\scriptsize full}}}} 

\newcommand{\meter}{\mbox{m}}
\newcommand{\millimeter}{\mbox{mm}}
\newcommand{\micron}{\mbox{$\mu$m}}
\newcommand{\nanometer}{\mbox{nm}}
\newcommand{\second}{\mbox{s}}
\newcommand{\millisecond}{\mbox{ms}}

\newcommand{\molecule}{\mbox{molecule}}

\newcommand{\kg}{\mbox{kg}}
\newcommand{\kiloWatt}{\mbox{kW}}
\newcommand{\picoWatt}{\mbox{pW}}
\newcommand{\Watt}{\mbox{W}}
\newcommand{\Joule}{\mbox{J}}
\newcommand{\Kelvin}{\mbox{K}}
\newcommand{\Pascal}{\mbox{Pa}}
\newcommand{\picoNewton}{\mbox{pN}}

\begin{document}

\title{Chemical Power for Microscopic Robots in Capillaries}
\author{Tad Hogg\\Hewlett-Packard Laboratories\\Palo Alto, CA \and Robert A.~Freitas Jr.\\Institute for Molecular Manufacturing\\Palo Alto,
CA}

\maketitle

\begin{abstract}

The power available to microscopic robots (nanorobots) that oxidize
bloodstream glucose while aggregated in circumferential rings on
capillary walls is evaluated with a numerical model using axial
symmetry and time-averaged release of oxygen from passing red blood
cells. Robots about one micron in size can produce up to several
tens of picowatts, in steady-state, if they fully use oxygen
reaching their surface from the blood plasma. Robots with pumps and
tanks for onboard oxygen storage could collect oxygen to support
burst power demands two to three orders of magnitude larger. We
evaluate effects of oxygen depletion and local heating on
surrounding tissue. These results give the power constraints when
robots rely entirely on ambient available oxygen and identify
aspects of the robot design significantly affecting available power.
More generally, our numerical model provides an approach to
evaluating robot design choices for nanomedicine treatments in and
near capillaries.

\textbf{Keywords:} nanomedicine, nanorobotics, capillary, power,
numerical model, oxygen transport
\end{abstract}

\newpage

\section{Introduction}

Nanotechnology has the potential to revolutionize health
care~\cite{morris01,nih03,keszler01,monroe09}. A current example is
enhanced imaging with nanoscale
particles~\cite{vodinh06,popovtzer08}. Future possibilities include
programmable machines comparable in size to
cells~\cite{freitas98,freitas99,freitas06,martel07}. Such
microscopic robots (``nanorobots'') could provide significant
medical benefits~\cite{freitas99,morris01,hill08}.

Realizing these benefits requires fabricating the robots cheaply and
in large numbers. Such fabrication is beyond current technology, but
could result from ongoing progress in developing nanoscale devices.
One approach is engineering biological systems, e.g., RNA-based
logic inside cells~\cite{win08} and bacteria attached to
nanoparticles~\cite{martel08}. However, biological organisms have
limited material properties and computational speed. Instead, we
consider machines based on plausible extensions of currently
demonstrated nanoscale electronics, sensors and
motors~\cite{barreiro08,berna05,collier99,craighead00,howard97,fritz00,marden02,montemagno99,wang05}
and relying on directed assembly~\cite{kufer08}. These components
enable nonbiological robots that are stronger, faster and more
flexibly programmed than is possible with biological organisms.

A major challenge for nanorobots arises from the physics of their
microenvironments, which differ in several significant respects from
today's larger robots. First, the robots will often operate in
fluids containing many moving objects, such as cells, dominated by
viscous forces. Second, thermal noise is a significant source of
sensor error and Brownian motion limits the ability to follow
precisely specified paths.
Finally, power significantly constrains the
robots~\cite{mallouk09,soong00}, especially for long-term
applications where robots may passively monitor for specific rare
conditions (e.g., injury or infection) and must respond rapidly when
those conditions occur.

Individual robots moving passively with the circulation can approach
within a few cell diameters of most tissue cells of the body. To
enable passing through even the smallest vessels, the robots must be
at most a few microns in diameter.  This small size limits the
capabilities of individual robots. For tasks requiring greater
capabilities, robots could form aggregates by using self-assembly
protocols~\cite{freitas99}. For robots reaching tissues through the
circulation, the simplest aggregates are formed on the inner wall of
the vessel. Robots could also aggregate in tissue spaces outside
small blood vessels by exiting capillaries via
diapedesis~\cite{freitas99}, a process similar to that used by
immune cells~\cite{ager03}.

Aggregates of robots in one location for an extended period of time
could be useful in a variety of tasks. For example, they could
improve diagnosis by combining multiple measurements of
chemicals~\cite{hogg06b}. Using these measurements, the aggregate
could give precise temporal and spatial control of drug
release~\cite{freitas99,freitas06} as an extension of an \textit{in
vitro} demonstration using DNA computers~\cite{benenson04}. Using
chemical signals, the robots could affect behavior of nearby tissue
cells. For such communication, molecules on the robot's surface
could mimic existing signalling molecules to bind to receptors on
the cell surface~\cite{freitas99,freitas03}. Examples include
activating nerve cells~\cite{vu05} and initiating immune
response~\cite{freitas03}, which could in turn amplify the actions
of robots by recruiting cells to aid in the treatment. Such actions
would be a small-scale analog of robots affecting self-organized
behavior of groups of organisms~\cite{halloy07}.
Aggregates could also monitor processes that take place over long
periods of time, such as electrical activity (e.g., from nearby
nerve cells), thereby extending capabilities of devices tethered to
nanowires introduced through the circulatory system~\cite{llinas05}.
In these cases, the robots will likely need to remain on station for
tens of minutes to a few hours or even longer.

The aggregate itself could be part of the treatment by providing
structural support, e.g., in rapid response to injured blood
vessels~\cite{freitas00}. Aggregates could perform precise
microsurgery at the scale of individual cells, extending surgical
capabilities of simpler nanoscale devices~\cite{leary06}. Since
biological processes often involve activities at molecular, cell,
tissue and organ levels, such microsurgery could complement
conventional surgery at larger scales. For instance, a few
millimeter-scale manipulators, built from micromachine (MEMS)
technology, and a population of microscopic devices could act
simultaneously at tissue and cellular size scales, e.g., for nerve
repair~\cite{sretavan05,hogg05}.

For medical tasks of limited duration, onboard fuel created during
robot manufacture could suffice. Otherwise, the robots need energy
drawn from their environment, such as converting externally
generated vibrations to electricity~\cite{wang07} or chemical
generators~\cite{freitas99}. Power and a coarse level of control can
be combined by using an external source, e.g., light, to activate
chemicals in the fluid to power the machines in specific
locations~\cite{tucker08}, similar to nanoparticle activation during
photodynamic therapy~\cite{bechet08}, or by using localized thermal,
acoustic or chemical demarcation~\cite{freitas99}.

This paper examines generating power for long-term robot activity
from reacting glucose and oxygen, which are both available in the
blood. Such a power source is analogous to bacteria-based fuel cells
whose enzymes enable full oxidation of
glucose~\cite{chaudhuri03,logan05,malki08}. We describe a
computationally feasible model incorporating aspects of
microenvironments with significant effect on robot performance but
not previously considered in robot designs, e.g., kinetic time
constants determining how rapidly chemical concentrations adjust to
robot operations. As a specific scenario, we focus on modest numbers
of robots aggregated in capillaries.

A second question we consider is how the robots affect surrounding
tissue. Locally, the robots compete for oxygen with the tissue and
also physically block diffusion out of the capillary. Robot power
generation results in waste heat, which could locally heat the
tissue. The robot oxygen consumption could also have longer range
effects by depleting oxygen carried in passing red blood cells.

In the remainder of this paper, we present a model of the key
physical properties relevant to power generation for robots using
oxygen and glucose in the blood plasma. Using this model, we then
evaluate the steady-state power generation capabilities of
aggregated robots and how they influence surrounding tissue.


\section{Modeling Physical Processes for Microscopic
Robots}\sectlabel{model}

We consider microscopic robots using oxygen and glucose available in
blood plasma as the robots' power source. This scenario involves
fluid flow, chemical diffusion, power generation from reacting
chemicals and waste heat production.
Except for the simplest geometries, behaviors must be computed
numerically, e.g., via the finite element method~\cite{strang73}.
Computational feasibility requires a choice between level of detail
of modeling individual devices and the scale of the simulation, both
in number of devices and physical size of the environment
considered. For microscopic biological environments relevant for
nanorobots, detailed physical properties may not be known or
measurable with current technology, thereby limiting the level of
detail possible to specify.

This section describes our model. The simplifying approximations are
similar to those used in biophysical models of microscopic
environments, such as oxygen transport in small blood vessels with
diffusion into surrounding tissue~\cite{mcguire01,popel89}. We focus
on steady-state behavior indicating long-term robot performance when
averaged over short-term changes in the local environment such as
individual blood cells (exclusively erythrocytes, not white cells or
platelets unless noted otherwise) passing the robots.

\subsection{Blood Vessel and Robot Geometry}\sectlabel{geometry}

\begin{figure}[t]
\begin{center}
\begin{tabular}{cc}
\includegraphics[width=\figwidthS]{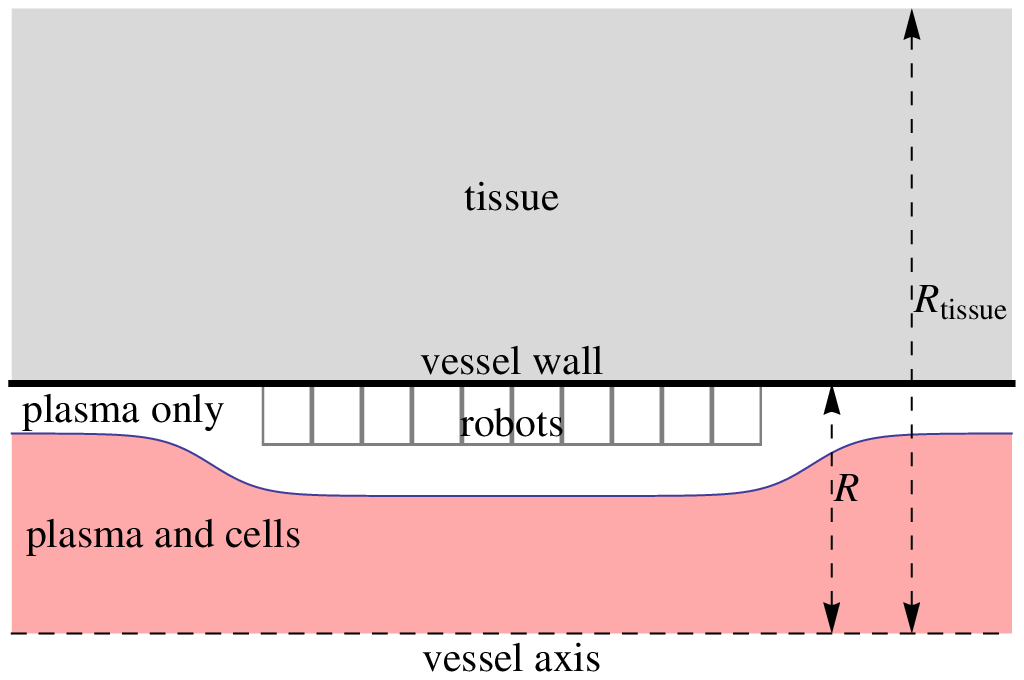} &
\includegraphics[width=\figwidthS]{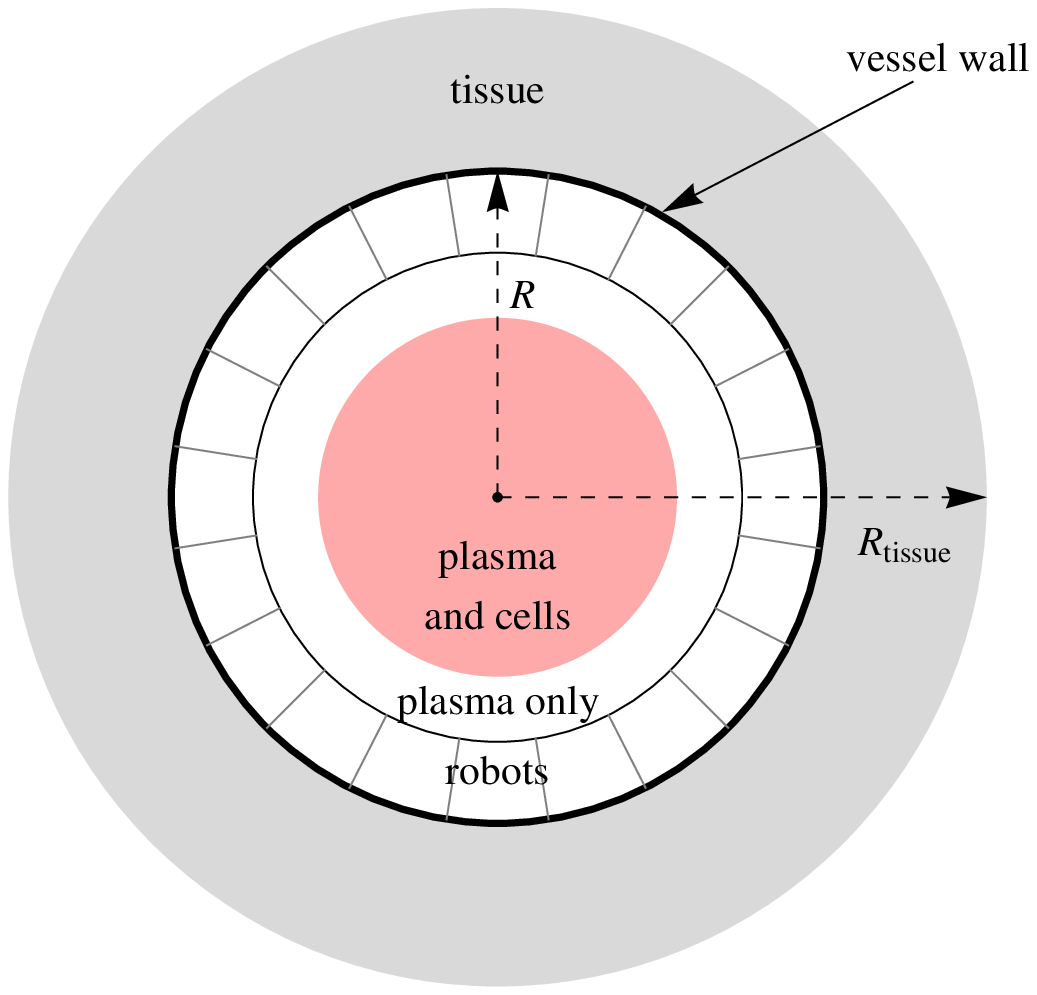} \\
(a)&(b)\\
\end{tabular}
\end{center}
\caption{\figlabel{geometry}Schematic geometry of vessel, robots and
surrounding tissue. The relative sizes of the regions are not to
scale for the parameters of our model (described in
\sect{parameters}). (a) A slice through the axially symmetric
geometry with the vessel axis at the bottom, showing a cross section
of 10 rings of robots. Fluid flows through the vessel from left to
right, with varying speed depending on distance to the vessel wall.
(b) Vessel cross section at the position of one ring of robots.}
\end{figure}

Evaluating behavior in general three-dimensional geometries is
computationally intensive. Simplified physical models give useful
insight with significantly reduced computational
requirements~\cite{popel89}. Such simplifications include using
two-dimensional and axially-symmetric three-dimensional geometries.
The latter case, appropriate for behavior within vessels, has
physical properties independent of angle of rotation around the
vessel axis. We adopt this approach and consider the
axially-symmetric geometry illustrated in \fig{geometry}: a segment
of a small vessel with robots forming one or more rings around the
vessel wall. The figure includes the Fahraeus effect: confinement of
blood cells near the center of the vessel. \sect{cells} describes
how we model this effect.

To ensure axial symmetry, we model the robot's interior with uniform
physical properties and take their shapes conforming to the vessel
wall with no gaps between neighboring robots~\cite{freitas99}. Thus,
as seen in \fig{geometry}(b), the surfaces of the robots contacting
the plasma or the vessel wall are curved, so the robots are only
roughly cubical. The other robot surfaces indicated in
\fig{geometry} are not treated explicitly in our model.

Physically uniform robot interiors are convenient but not necessary
for this model. An axially symmetric model only requires the robots
to be uniform in the direction around the vessel and that the radial
boundary surfaces between robots are treated as continuous with the
interiors. Robot characteristics could vary in the direction along
the vessel axis or radially. For example, the axially symmetric
model could apply to robots whose power generators are close to the
plasma-contacting surface to minimize internal oxygen transport,
analogous to clumping of mitochondria in cells near
capillaries~\cite{popel89}. Moreover, while we primarily focus on
physically adjacent rings of robots, axial symmetry also holds for
sets of rings that are spaced apart from each other along the
vessel. We refer to a set of rings of robots as a \emph{ringset}.

We ignore pulsatile variations in vessel circumference as these are
mostly confined to the larger arterial vessels~\cite{silver87}. Thus
our model geometry is both axially symmetric and static.

\subsection{Fluid Flow}

Viscosity dominates the motion of microscopic objects in fluids,
producing different physical behaviors than for larger organisms and
robots in
fluids~\cite{purcell77,vogel94,fung97,karniadakis05,squires05}.
The Navier-Stokes equation describes the
flow~\cite{fetter80,karniadakis05}. For the vessel geometry of
\fig{geometry}, the pressure difference between the inlet and outlet
of the vessel determines the nature of the flow. We specify the
pressure difference as $\pressureGradient L$ where
$\pressureGradient$ is the overall pressure gradient and $L$ is the
length of the modeled segment of the vessel. While some fluid leaks
into or out of capillaries, we ignore this small component of the
flow, in common with other models of blood flow in
capillaries~\cite{popel89,pozrikidis05}. In our scenario, the robots
are attached to the vessel wall. For modeling fluid behavior, such
static robots merely change the shape of the vessel boundary. We
apply the ``no slip'' boundary condition on both the robots and the
vessel wall, i.e., fluid speed is zero at these boundaries. Thus the
flow speed varies from zero at the wall to a maximum value in the
center of the vessel.

\subsection{Chemical Diffusion}

Microscopic robots and bacteria face similar physical constraints in
obtaining chemicals~\cite{berg77}. At small scales, diffusion
arising from random thermal motions is the main process transporting
chemicals. Even at the scale of these robots, individual molecules
and their distances between successive collisions are tiny. Thus
chemicals in the fluid are well-approximated by a continuous
concentration $C$ specifying the number of molecules per unit
volume. The concentration obeys the diffusion
equation~\cite{berg93},
\begin{equation}\eqlabel{diffusion}
\frac{\partial C}{\partial t} = -\nabla \cdot \Flux + \Gamma
\end{equation}
where $\Flux = -D \nabla C + \vFluid C$ is the chemical flux,
$\nabla C$ is the concentration gradient, $\vFluid$ is the fluid
velocity vector, $D$ is the chemical diffusion coefficient and
$\Gamma$ is the reaction rate density, i.e., rate at which molecules
are created by chemical reactions per unit volume. The first term in
the flux is diffusion, which acts to reduce concentration gradients,
and the second term arises from movement of the fluid in which the
chemical is dissolved.

Small molecules such as oxygen and glucose readily diffuse from
capillaries into surrounding tissue. \eq{diffusion} also describes
the transport within the tissues wherein $\vFluid \approx 0$, i.e.,
the transport is completely due to diffusion. The diffusion
coefficient of oxygen in tissue is close to that in
plasma~\cite{mcguire01}, and for simplicity we use the same
diffusion coefficients in both regions.

\subsection{Kinetics of Oxygen Release from Red Blood
Cells}\sectlabel{kinetics}

As robots consume oxygen from the plasma, passing red blood cells
respond to the reduced concentration by releasing oxygen. An
important issue for powering robots is how rapidly cells replenish
the oxygen in the plasma as the cells pass the robots.

A key value determining the oxygen release from red blood cells is
the hemoglobin saturation $S$: the fraction of hemoglobin capacity
in a cell which has bound oxygen~\cite{mauroy07}. The oxygen
concentration in the cell is $\CoxygenMax S$, where $\CoxygenMax$ is
the concentration in the cell when all the hemoglobin has bound
oxygen.

The saturation is high when the cell is in fluid with high oxygen
content, i.e., in the lungs, and low after the cell has delivered
oxygen to tissues of the body. Quantitatively, the equilibrium
saturation, conventionally expressed in terms of the equivalent
partial pressure $\pressure$ of \Oxygen\ in the fluid around the
cell, is well-described by the Hill equation~\cite{popel89}:
\begin{equation}\eqlabel{equilibrium saturation}
\Sequib(a) = \frac{a^\nHill}{1+a^\nHill}
\end{equation}
where $a=\pressure/\Phalf$ is the partial pressure ratio, $\Phalf$
is the partial pressure at which half the hemoglobin is bound to
oxygen and $\nHill$ characterizes the steepness of the change from
low to high saturation. The saturation in small blood vessels ranges
from near $1$ within the lungs to around $1/3$ within working
tissues. Henry's Law relates the partial pressure to the
concentration: $p = \Hoxygen \Coxygen$ with the proportionality
constant $\Hoxygen$ depending on the fluid temperature.

\eq{equilibrium saturation} gives the equilibrium saturation, i.e.,
the value in a red cell after residing a sufficiently long time in a
fluid with partial pressure $p$. However, small robots consuming
oxygen from the plasma may produce large gradients in oxygen
concentration. If the oxygen concentration gradients and flow speed
are high enough, passing cells will not have time to equilibrate
with the abruptly decreased oxygen concentration before the flow
moves them past the robots. Whether this is the case depends on the
kinetics, i.e., how rapidly cells change their saturation level when
exposed to concentration changes. The time scale for oxygen release
is determined by reaction kinetics of oxygen binding to hemoglobin
in the cell and diffusion of these chemicals within the cell.

One model of this kinetics is a lumped-model differential equation
relating saturation to concentration outside the
cell~\cite{clark85}. In this model, the change in $S$, and hence the
flux of oxygen from a cell into the surrounding plasma, is
determined from the partial pressure ratio $a$ as
\begin{equation}\eqlabel{S(t)}
\frac{d S(t)}{d t} = -\frac{1}{\tUnloading} \sqrt{s(a,S(t))}
\end{equation}
where $\tUnloading$ is a characteristic time scale for oxygen
unloading and the saturation unloading function $s$ is
\begin{equation}\eqlabel{saturation}
s(a,S) = \frac{2(1-S)}{\nHill+1} a^{\nHill+1} - 2 S a +
\frac{2\nHill}{\nHill+1} \frac{S^{1+1/\nHill}}{(1-S)^{1/\nHill}}
\end{equation}
If oxygen partial pressure, $\pressure$, varies over the surface of
the cell, the rate of change is the average of the right-hand-side
of \eq{S(t)} over the surface of the cell. \eq{S(t)} is consistent
with the equilibrium relation of \eq{equilibrium saturation} because
$s(a,\Sequib(a))=0$.

As a boundary condition on oxygen saturation $S$, at the vessel
inlet we take $S$ equal to the equilibrium value with the oxygen
plasma concentration specified at the inlet. Numerical evaluation of
\eq{S(t)} requires care to accurately evaluate $s$ when the
concentration is close to equilibrium to avoid numerical instability
if the computed concentration in the plasma is even slightly above
the cell saturation.

The blood cells also have a role in removing the carbon dioxide
produced by the robots (\sect{robot power}). Only a small portion is
transported dissolved in the plasma. Instead, most \CarbonDioxide\
is transported or chemically converted to bicarbonate within red
cells. The detailed kinetics of these processes~\cite{geers00} does
not directly limit robot power production, and thus is beyond the
scope of this paper. Moreover, the robot power production rates
considered here increase the carbon dioxide concentration by only a
few percent, which can be buffered by processes within the passing
cells and so is not likely to be a safety constraint on the power
levels in the scenarios we consider.

\subsection{Robot Power Generation}\sectlabel{robot power}

The overall chemical reaction combining
glucose and oxygen to produce water and carbon dioxide is
\begin{displaymath}
\mbox{C}_6\mbox{H}_{12}\mbox{O}_6 + 6\,\Oxygen \rightarrow
6\,\CarbonDioxide + 6\,\mbox{H}_2\mbox{O}
\end{displaymath}
We denote the energy released by each such reaction by
$\reactionEnergy$. A robot absorbing oxygen molecules at a rate
\totalOxygenFlux\ produces power $\totalOxygenFlux
\reactionEnergy/6$ because each reaction uses six \Oxygen\ molecules.

We consider robots on the vessel wall absorbing chemicals from the
fluid only on their plasma-facing sides. For generating power with
oxygen and glucose from the blood, oxygen is the limiting
chemical~\cite{freitas99}. We examine two design choices for the
robots: how they collect oxygen arriving at their surface and their
capacity for processing that oxygen to produce power.

For the first design choice, oxygen transport within the robots, we
examine two extremes. In the basic (``no pumps'') design, the robots
absorb oxygen passively via diffusion. In the advanced (``with
pumps'') design, the robots use pumps on their surfaces to actively
absorb all arriving oxygen and distribute this gas to internal power
generating sites.
We treat the full surface as available to absorb chemicals. In
practice, robots will absorb chemicals with only a fraction of their
surface. This is not a significant constraint for microscopic robots
since even a modest fraction of a surface with absorbing sites gives
absorption almost as large as that of a fully absorbing
surface~\cite{berg77}.

For the second design choice, robot power production capacity, we
also examine two cases. Onboard generating capacity arises from the
number and efficiency of the internal reaction sites, e.g., fuel
cells~\cite{freitas99,chaudhuri03,malki08}, in each robot. If
capacity is constrained by engineering feasibility of fuel cell
fabrication or by difficulty of placement into the robots, the
robots will have relatively few fuel cells -- and consequently a low
maximum capacity for power generation -- hence are called ``low
capacity'' robots. When these constraints do not apply, we have
``high capacity'' robots.

A robot with sufficient pump and generating capacity produces power
from all oxygen reaching the robot. This oxygen-limited situation
corresponds to a zero-concentration boundary condition for the
oxygen concentration in the fluid at the robot surface. With this
boundary condition, integrating the dot product of the flux \Flux\
(determined from \eq{diffusion}) and normal vector of the
plasma-facing surface of the robot gives the rate \totalOxygenFlux\
(molecules per unit time) at which the robot absorbs oxygen
molecules, with no need to explicitly model oxygen transport and
consumption inside the robot.
While pumps cannot maintain the zero-concentration boundary
condition at arbitrarily high oxygen flux, theoretical pump capacity
appears more than adequate for the oxygen concentrations relevant to
our model~\cite{freitas99}.

A robot's power generating capacity is limited by the number of
reaction sites it contains, \reactionSiteNumber, and by the maximum
rate of reacting glucose and oxygen at each site, \reactionRate.
Specifically, the steady-state oxygen absorption rate must satisfy
$\totalOxygenFlux \leq 6 \reactionSiteNumber \reactionRate$. If this
bound on absorption rate is smaller than the oxygen flux
corresponding to the zero-concentration boundary condition the pumps
could maintain, then the robot's power generation will be
capacity-limited rather than oxygen-limited and the
zero-concentration boundary condition will not apply. In this
situation, for a robot with pumps we consider the pumps delivering
as much oxygen as the reaction sites can process, giving robot power
generation equal to its maximum possible value, namely,
$\reactionSiteNumber \reactionRate \reactionEnergy$.

Determining power generation for robots without pumps requires
explicitly modeling the oxygen transport and power generation within
the robot. In this case the oxygen moves by diffusion within the
robot. We treat power generation as spatially continuous rather than
occurring at discrete reaction sites, thereby maintaining axial
symmetry. Thus \eq{diffusion} applies within the robot, with the
reaction rate density for oxygen, $\Gamma$, determined by the number
density of reaction sites, $\reactionSiteDensity$, and the reaction
kinetics of each site. For uniformly distributed reaction sites,
$\reactionSiteDensity = \reactionSiteNumber/\Vrobot$ where $\Vrobot$
is the volume of each robot. Specifically, at a given location
inside the robot, $\Gamma = 6 \PowerRobot/\reactionEnergy$ where
$\PowerRobot$ is the power generation density, i.e., the power
generated per unit volume at that location.
We model robot power generation using Michaelis-Menten
kinetics~\cite{briggs25} assuming oxygen is the limiting factor
because glucose concentrations are typically two orders of magnitude
larger than those of oxygen~\cite{freitas99}:
\begin{equation}\eqlabel{reaction rate}
\PowerRobot = \reactionEnergy \reactionSiteDensity \reactionRate
\frac{\Coxygen}{\ChalfReaction + \Coxygen}
\end{equation}
where $\ChalfReaction$ is the concentration of $\Oxygen$ giving half
the maximum reaction rate.
The total power generated by a robot is the integral of
$\PowerRobot$ over the robot's volume, which is the same as the
power determined from the rate the robot absorbs oxygen, i.e.,
$\reactionEnergy\totalOxygenFlux /6$. In this no-pumps case
$\totalOxygenFlux$ is determined from the solution of \eq{diffusion}
in the fluid and robot interior rather than from a boundary
condition on the robot's surface.

The number of reaction sites in a robot is a design choice, limited
by the volume of each reaction site. As an example, a nanoscale
oxygen-glucose fuel cell could be as small as $3000\,\nanometer^3$
with $\reactionRate = 10^6$ glucose molecules per
second~\cite{freitas99}. $\reactionSiteDensity$ can't be larger than
the reciprocal of this volume -- which would correspond to the robot
entirely filled by power generation reaction sites. To illustrate
the tradeoffs among these design choices, we consider high and low
capacity robots, both with and without pumps. Increasing oxygen
concentration at the reaction sites increases their power output
closer to their maximum (since the fraction appearing in
\eq{reaction rate} gets closer to its maximum value of one). Thus
pumps can at least somewhat compensate for a decrease in the number
of functional reaction sites by increasing the oxygen concentration
so the remaining reaction sites operate more efficiently. On the
other hand, if pumps are more difficult to fabricate than fuel
cells, robots would benefit from a large number of fuel cells (high
capacity) to compensate for the inability of passive diffusion to
increase concentrations. As another approach to dealing with few
fuel cells, we also consider placing all of them near the
plasma-facing surface of the robot, where oxygen concentration is
highest in the passive diffusion (no-pumps) design.

\subsection{Oxygen Use in Tissues}

Models of oxygen use and power generation in tissues can include
various details of tissue structure~\cite{popel89}. A simple
approach, adopted in this paper, treats the tissue surrounding the
vessel as homogeneous and metabolizing oxygen (assumed to be the
rate-limiting chemical) with kinetics similar in form to
\eq{reaction rate}:
\begin{equation}\eqlabel{tissue reaction rate}
\PowerTissue = \PowerTissueMax \frac{\Coxygen}{\ChalfReactionTissue
+ \Coxygen}
\end{equation}
where $\PowerTissueMax$ is the power demand (power per unit volume)
of the tissue and $\ChalfReactionTissue$ is the concentration of
$\Oxygen$ giving half the maximum reaction rate.

\subsection{Heating}

The robot-generated power eventually dissipates as waste heat into
the environment. Heat transfer from the robots to their surroundings
occurs by both conduction and convection due to the moving fluid. We
take the tissue environment outside the vessel to be small enough so
as not to include other vessels. Thus heat transport in the tissue
is via conduction only.

The temperature $T$ obeys a version of
\eq{diffusion}~\cite{fetter80}:
\begin{equation}\eqlabel{heat}
\density \, \heatCapacity \frac{\partial T}{\partial t} = -\nabla
\cdot \Flux + Q
\end{equation}
where $\Flux = -\kThermal \nabla T + \density \, \heatCapacity T
\vFluid$ is the heat flux, $\nabla T$ is the temperature gradient,
$\vFluid$ is the fluid velocity vector, $\rho$ is fluid density,
$\kThermal$ is the fluid's thermal conductivity, $\heatCapacity$ is
the fluid's heat capacity, and $Q$ is the heat generation rate
density which is the same as the power production per unit volume.
For robots absorbing all oxygen reaching them (i.e., using pumps),
we take $Q$ uniform within the robot, i.e., equal to
$\reactionEnergy\totalOxygenFlux/(6 \Vrobot)$. For robots without
pumps, power generation varies within the robot, with
$Q=\PowerRobot$ from \eq{reaction rate}. For temperature boundary
conditions, we take the incoming fluid and the outermost radius of
the tissue cylinder to be held at body temperature.

While we could include tissue power generation as a heat source in
the heat equation, here we focus on the \emph{additional} heat from
the robots alone. Thus we evaluate how robot power generation adds
to the heat load produced by the tissue. We do not consider any
changes in the tissue, either locally or systemically (e.g.,
increasing blood flow), in response to the additional heating. This
is a reasonable assumption given the tiny temperature increase
described in \sect{tissue power and heating}.

\subsection{Effects of Cells on Flow and Chemical Transport}\sectlabel{cells}

In small blood vessels, individual blood cells are comparable in
size to the vessel diameter. Thus, at the length scales relevant for
microscopic robots, the fluid consists of plasma separating
relatively large objects. The cells significantly affect the fluid
flow and, because cells are not rigid, the flow alters the shape of
the cells (though we can ignore red blood cell rotation-induced
elevation of diffusivity~\cite{keller71} because these cells are
motionally restricted in capillaries and elevation is lowest for
small molecules such as \Oxygen). Similarly, the vessel walls are
not rigid, which somewhat changes both the flow and the vessel
boundary.
A key consequence for oxygen transport is the confinement of cells
toward the center of the vessel. The cell-free fluid near the vessel
wall is a gap over which oxygen released by cells must diffuse to
reach the vessel wall or the plasma-facing robot surface. In
capillaries, this gap ranges from about 1/2 to 1 micron, depending
on flow speed~\cite{secomb01,pozrikidis05}.

Modeling the interactions between fluid and blood cells is
computationally feasible for a few cells in
capillaries~\cite{mauroy07,hogg08h}. However, modeling interactions with
many deforming cells is challenging and close packing of objects
moving in fluid leads to complex hydrodynamic
interactions~\cite{hernandez05,riedel05}. Instead of evaluating
these effects in detail, we use approximate models that average over
the cell behaviors and assume rigid vessel walls. Such models are
commonly used to study oxygen delivery in tissue~\cite{popel89}.
This averaging approach also simplifies analysis of collective robot
behavior~\cite{lerman01,hogg06a,hamann08}.

In this approximation, the vessel only contains fluid, which
consists of two components as illustrated in \fig{geometry}. The
first component models the mix of cells and plasma in the central
portion of the vessel. Instead of explicitly modeling individual
cells this approximation averages over the cell positions in the
fluid. The second component is the fluid near the vessel wall,
consisting of plasma only.

The fluid component modeling the mix of cells and plasma is confined
to a distance $\Rcell$ from the vessel axis. This distance varies
with position along the vessel, as shown in \fig{geometry}, since
robots on the wall reduce the volume available to the passing fluid.
Thus all oxygen released by the passing cells is within a distance
$\Rcell$ of the vessel axis, and this oxygen must diffuse through
the plasma gap to reach the robots or the tissue. We take $\Rcell$
to follow a fluid streamline with the gap appropriate for the fluid
speed in the section of the vessel far from the
robots~\cite{secomb01}. This approximation accounts for the location
of cells toward the center of the vessel without the complexity of
modeling how cells change shape as they pass the robots.

A key parameter for oxygen delivery is the hematocrit,
\hematocritFull, i.e., the fraction of the capillary volume occupied
by cells. In our model, the more relevant parameter is the
hematocrit, \hematocrit, within the fluid component containing the
cells, which has a smaller volume than the full vessel. Since both
values must give the same rate for cells passing through the vessel,
these quantities are related by
\begin{equation}\eqlabel{hematocrit}
\hematocrit = \hematocritFull \frac{R^2 \vAvg}{\Rcell^2 \vAvgCell}
\end{equation}
where $\vAvg$ is the average flow speed in the vessel and
$\vAvgCell$ is the average flow speed within the central portion of
the vessel with fluid component modeling the cells. Fluid speed is
faster near the center of the vessel than near the walls, so
$\vAvgCell$ is larger than $\vAvg$. The quantities $\vAvg$,
$\vAvgCell$ and $\Rcell$ vary along the length of the vessel, but
the ratio appearing in \eq{hematocrit} is constant due to our choice
of $\Rcell$ following a fluid flow streamline.
Within the cell fluid component, oxygen bound to hemoglobin has
concentration $\hematocrit \CoxygenMax S$ and oxygen in the plasma
has concentration $(1-\hematocrit)\Coxygen$. Future evaluations of
the accuracy of this simplifying approach to oxygen delivery might
include results from more detailed models comparing oxygen release
from red cells with that of hemoglobin-based oxygen carriers
dissolved in plasma rather than contained in
cells~\cite{vadapalli02}. This averaging over cell position can also
be viewed as approximating the time-averaged behavior as cells pass
the robots on the vessel wall.

We model the kinetics of oxygen release from passing cells as due to
changes in cell saturation in the cell fluid component, i.e., $S$.
The effect of oxygen release from red cells into the plasma arises
from the rate of change in saturation inside the
cells~\cite{clark85}, as discussed in \sect{kinetics} with
\eq{S(t)}. Thus the reaction term in \eq{diffusion} for oxygen in
the fluid component with the cells is
\begin{equation}\eqlabel{oxygen reaction rate}
\Gamma = -h \CoxygenMax \; \frac{d S}{d t}
\end{equation}
Since $dS/dt$ from \eq{S(t)} is negative, this
value for $\Gamma$ is positive, giving an increase in oxygen in the
plasma.

We determine $S$ along the vessel using the lumped model discussed
in \sect{kinetics}. The value of $S$ along the vessel is governed by
a one-dimensional version of the diffusion equation based on the
average flow speed in the cell fluid component, $\vAvgCell$, and
using the chemical diffusion coefficient for oxygen bound to
hemoglobin in the cell, $\Dheme$. We determine the reaction term in
the diffusion equation for $S$, for each position along the vessel,
by averaging the right-hand side of \eq{S(t)} over the cross section
of the vessel at that position, based on the oxygen concentration in
the plasma of the plasma and cell component of the fluid. This
average value gives the rate of change for the saturation of cells
as they pass that position along the vessel. In this way the changes
in saturation within the cells and the concentration in the plasma
are coupled equations that are solved simultaneously.

\subsection{Model Parameters}\sectlabel{parameters}

\begin{table}[t]
\begin{center}
\begin{tabular}{lc}
parameter    &   value \\ \hline
\hline \multicolumn{2}{c}{\bf geometry} \\
vessel radius    &   $R=4\,\micron$   \\
tissue cylinder radius   & $\Rtissue=40\,\micron$ \\
modeled vessel length  &   $L=100\,\micron$    \\
\hline \multicolumn{2}{c}{\bf fluid} \\
ambient temperature    &   $T=310\,\Kelvin$   \\
thermal conductivity    &   $\kThermal = 0.6\,\Watt/\meter/\Kelvin$
\\
heat capacity    &   $\heatCapacity = 4200\,\Joule/\kg/\Kelvin$   \\
fluid density   & $\density=10^3 \,\kg/\meter^3$ \\
fluid viscosity & $\viscosity=10^{-3} \,\kg/\meter/\second$ \\
\textbf{pressure gradient}  &   $\pressureGradient=1 \mbox{--}
5\times 10^5\,\Pascal/\meter$ \\ 
hematocrit    &   $\hematocritFull = 25\%$   \\
\hline \multicolumn{2}{c}{\bf tissue} \\
\textbf{power demand}    & $\PowerTissueMax = 4\mbox{--}60 \,\kiloWatt/\meter^3$ \\
\Oxygen\ concentration for half power   & $\ChalfReactionTissue = 10^{21}\, \molecule/\meter^3 $ \\
reaction energy from one glucose molecule  & $\reactionEnergy =
4\times
10^{-18}\,\Joule$\\
density, thermal conductivity, heat capacity & same as fluid\\
\hline \multicolumn{2}{c}{\bf red blood cells} \\
partial pressure for 50\% $\Oxygen$ saturation   & $\Phalf =
3500\,\Pascal$ \\
$\Oxygen$ saturation exponent    & $\nHill = 2.7$ \\
time constant for $\Oxygen$ unloading   & $\tUnloading = 76\,\millisecond$ \\
maximum $\Oxygen$ concentration in cell  & $\CoxygenMax = 10^{25}\,
\molecule/\meter^3 $ \\
heme diffusion coefficient   & $ \Dheme = 1.4\times 10^{-11}\, \meter^2/\second$ \\
\hline \multicolumn{2}{c}{\bf chemicals in plasma} \\
$\Oxygen$ diffusion coefficient & $\Doxygen=2 \times 10^{-9}\,\meter^2/\second$ \\
\textbf{$\Oxygen$ concentration at inlet}   &  $\Coxygen = 3\mbox{--}7 \times 10^{22}\,\molecule/\meter^3$ \\ 
\Oxygen\ partial pressure to concentration ratio & $\Hoxygen =
1.6\times 10^{-19}\,\Pascal/(\molecule/\meter^3)$
\\
\end{tabular}
\end{center}
\caption{\tbllabel{parameters}Model parameters for fluid, vessel and
tissue. We consider two values, the extremes of the listed range,
for parameters indicated in bold. For the vessel without robots, the
pressure gradient range corresponds to average flow speeds of
$\vAvg=0.2 \mbox{--} 1\,\millimeter/\second$. The corresponding
hematocrit values within the cell fluid component are
$\hematocrit=0.31\mbox{--}0.36$. See text for source references.}
\end{table}

\tbl{parameters} lists the parameter values we use. To locate the
boundary between the fluid component modeling the cells and the
cell-free component near the vessel wall, we use cell-free gaps of
$0.98$ and $1.27\,\micron$ at the vessel inlet for pressure
gradients of $10^5$ and $5\times10^5\,\Pascal/\meter$, respectively.
For the $10\,\micron$ ringset, the fluid streamline becomes nearly
flat (i.e., fluid velocity in the radial direction is nearly zero)
near the middle of the aggregate, and the corresponding cell-free
gap for the narrow section of the vessel by the robots (i.e., of
radius $3\,\micron$) matches that for a long vessel with radius
$3\,\micron$~\cite{secomb01}.

We assume the fluid properties (i.e., density, viscosity, heat
capacity and thermal conductivity) are uniform throughout the model
and roughly equal to those of water. The pressure gradient range we
consider corresponds to average flow speeds of
$0.2\mbox{--}1\,\millimeter/\second$ in a vessel of radius $R$
without robots.
These speeds are typical of measured flow in
capillaries~\cite{freitas99}. For comparing vessels with and without
robots we use the same pressure gradients in both cases. That is, we
compare constant-pressure boundary conditions rather than
constant-velocity conditions.
The ambient temperature is body temperature and the hematocrit value
is typical of small blood vessels~\cite{freitas99}, which is
somewhat lower than in larger vessels.

For the kinetics, $\ChalfReactionTissue$ is from
Ref.~\cite{mcguire01} and the blood cell kinetics parameters are
from Refs.~\cite{clark85} and~\cite{popel89}.
The oxygen concentration range corresponds to venous and arterial
ends of capillaries~\cite{freitas99}. Concentrations of glucose and
$\CarbonDioxide$ in blood plasma are in the millimolar range (about
$10^{24}\,\molecule/\meter^3$), far larger than the oxygen
concentrations~\cite{freitas99}.
For evaluating microscopic robot behavior, a convenient measure of
chemical concentration in a fluid is number of molecules per unit
volume. Much of the existing literature uses units convenient for
larger scales, such as moles of chemical per liter of fluid (i.e.,
molar, M) and grams of chemical per cubic centimeter. Discussions of
gases dissolved in blood often specify concentration indirectly via
the corresponding partial pressure of the gas under standard
conditions. As an example, oxygen concentration $\Coxygen = 10^{22}
\,\molecule/\meter^3$ corresponds to a $17 \,\mbox{$\mu$M}$
solution, $0.53\,\mbox{$\mu$g}/\mbox{cm}^3$ and to a partial
pressure of $1600\,\Pascal$ or $12\,\mbox{mmHg}$.

Tissue power demands vary considerably, depending on the tissue type
and overall activity level. We consider typical values of resting
and high power demand~\cite{freitas99} and focus on two extreme
scenarios given in \tbl{scenarios}. The low demand scenario is the
likely situation for most medical procedures in practice. The high
demand scenario has a relatively high tissue demand, but is not the
peak metabolic rate in human tissue, which can reach rates as high
as $200\, \kiloWatt/\meter^3$~\cite{mcguire01}.

\begin{table}[t]
\begin{center}
\begin{tabular}{l|cc}
    & \multicolumn{2}{c}{scenario}\\
parameter & low demand & high demand \\
\hline pressure gradient & $\pressureGradient= 10^5\,\Pascal/\meter$
& $\pressureGradient=5\times 10^5\,\Pascal/\meter$ \rule{0pt}{10pt}\\
tissue power demand & $\PowerTissueMax = 4\,\kiloWatt/\meter^3$& $\PowerTissueMax = 60 \,\kiloWatt/\meter^3$\\
\end{tabular}
\end{center}
\caption{\tbllabel{scenarios}Two scenarios: \emph{low} with low
tissue power demand (basal rate) and slow fluid flow, and
\emph{high} with high tissue power demand and fast fluid flow. Both
scenarios use high inlet concentration: $\Coxygen = 7 \times
10^{22}\,\molecule/\meter^3$ (arterial).}
\end{table}

Fluid and chemical properties vary with temperature, but, as
described below, the temperature range seen in our model is very
small. Thus we take the values at body temperature. We also treat
the saturation curve of \eq{saturation} as constant although it
varies somewhat with \CarbonDioxide\ concentration through a change
in $\Phalf$.

\begin{table}[t]
\begin{center}
\begin{tabular}{lc}
parameter    &   value \\ \hline
\hline \multicolumn{2}{c}{\bf geometry} \\
robot size  & $\Rrobot = 1\,\micron$ \\
robots per circumferential ring & 20\\
robot volume & $\Vrobot = 1.1\,\micron^3$ \\
\textbf{length of aggregate} & $1\mbox{--}10\,\micron$ \\
\hline \multicolumn{2}{c}{\bf power generation}  \\ \textbf{power
generation site density}   & $\reactionSiteDensity =
0.06\mbox{--}3\times 10^{21}/\meter^3$
\\
power generation reaction rate  & $\reactionRate = 10^6/\second$ \\
\Oxygen\ concentration for half power   & $\ChalfReaction =
10^{24}\,\molecule/\meter^3$ \\
\end{tabular}
\end{center}
\caption{\tbllabel{robot parameters}Robot design parameters. The
robot size is the length of each robot in the radial and
longitudinal directions. The curved surfaces facing the plasma and
vessel wall have slightly different lengths. When not limited by
availability of oxygen or glucose, a power generation reaction site
produces $\reactionEnergy \reactionRate = 4\, \picoWatt$. We
consider two values, the extremes of the listed range, for
parameters indicated in bold. The number of power generation sites
in each robot, $\reactionSiteNumber = \reactionSiteDensity\Vrobot$,
ranges from $66$ to $3300$, with corresponding maximum power per
robot, $\reactionSiteNumber \reactionRate \reactionEnergy$, of
$260\,\picoWatt$ and $13000\,\picoWatt$ for low and high capacity
robots, respectively.}
\end{table}

The robot size, number aggregated on the vessel wall and power
generation capacity are design choices, with the values we consider
given in \tbl{robot parameters}. We consider sets of circumferential
rings along the vessel wall either one or ten adjacent robots long.
These aggregates consist of 20 and 200 robots, respectively.
We estimate $\ChalfReaction$ as the value corresponding to fuel
cells based on the glucose oxidase enzyme~\cite{bao01}. The high and
low capacity robot designs correspond to the choices of reaction
site density given in \tbl{robot parameters}. For the high capacity
case, the power generation uses about 1\% of the robot volume with
the fuel cells described in \sect{robot power}. As shown in
\sect{results} the maximum power generation, even for the low
capacity case, is considerably larger than possible with the
available oxygen. So these design choices are reasonable for
studying limitations due to available oxygen.

In our model, the fluid flow is independent of the chemical
concentrations, and both are independent of the heat generation due
to our assumption that the parameters of \tbl{parameters} are
independent of temperature in the narrow physiological range. This
simplifies the numerical solution by allowing an iterative
procedure: solving first for the fluid flow, then for the chemical
concentration and finally for the temperature.
Specifically, we first solve for the fluid flow in the vessel as
determined by the vessel and robot geometry and the imposed pressure
gradient. Given the fluid velocity, we then simultaneously solve
\eq{diffusion} for the oxygen concentration and \eq{S(t)} for the
blood cell average oxygen saturation. \eq{reaction rate} and
\eq{tissue reaction rate} give the power generation density in the
robots and tissue, respectively. For robots with pumps, we impose
the boundary conditions on the plasma-facing robot surface described
in \sect{robot power} and do not need to solve \eq{diffusion} inside
the robot. Dividing the power generation by $\reactionEnergy/6$,
where $e$ is the energy per reaction, gives the corresponding oxygen
reaction rate densities $\Gamma$ appearing in \eq{diffusion}, i.e,
the number of oxygen molecules consumed per unit volume per unit
time at each location. This solution gives the oxygen concentration
$\Coxygen$ and flux $\Flux$ throughout the vessel and the tissue,
and the average cell saturation $S$ as a function of distance along
the vessel. Finally, solving \eq{heat} using the solutions for the
fluid flow and power generated by the robots gives the temperature
increase due to the robots. We solve for steady-state behaviors,
though the model also applies to time-dependent scenarios.

For the numerical solution, we used a multiphysics finite element
solver~\cite{web.comsol} with about ten to twenty thousand mesh
points in the two-dimensional geometry representing a slice through
the axially symmetric geometry shown in \fig{geometry}a. We used the
default meshing procedure except constraining the mesh point spacing
along the plasma-facing robot surface to be at most $0.1\,\micron$
for the 10-micron ringset and $0.01\,\micron$ for the 1-micron ring.
This constraint gives tiny spacing between mesh nodes in the region
where the concentration is changing most rapidly, i.e., near the
robot surface. To ensure numerical stability when evaluating
\eq{S(t)}, if the plasma concentration is above the equilibrium
saturation of \eq{equilibrium saturation}, we use the opposite sign
in \eq{S(t)} so cell saturation increases rather than decreases.
This situation only occurs to a slight extent, due to numerical
errors in evaluating $S$ and the concentration in the plasma when
$S$ is close to the equilibrium value. We solve for the average cell
saturation as a function of position along the vessel in a
one-dimensional geometry with 900 mesh points.

\section{Results}\sectlabel{results}

\begin{figure}[t]
\begin{center}
\begin{tabular}{l|r}
\includegraphics[width=\figwidthS]{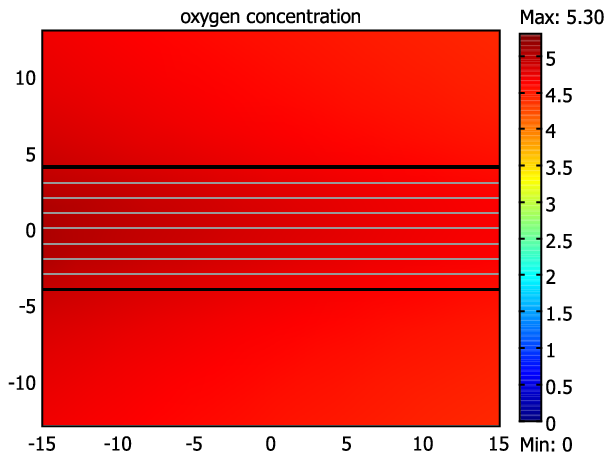} & 
\includegraphics[width=\figwidthS]{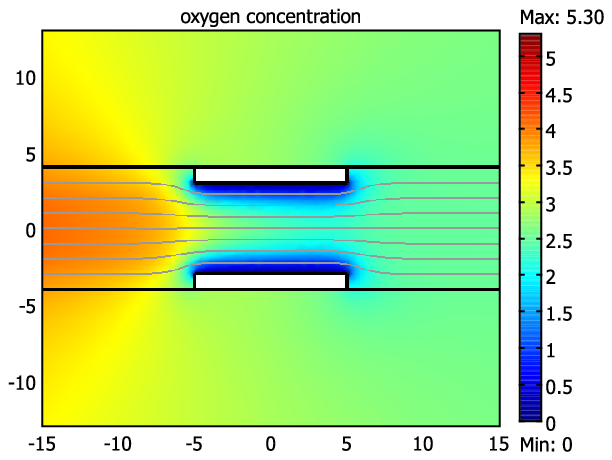}\\
\hline
\includegraphics[width=\figwidthS]{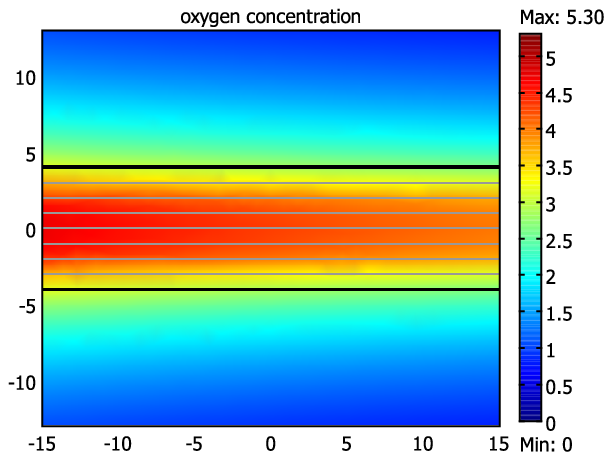}
&
\includegraphics[width=\figwidthS]{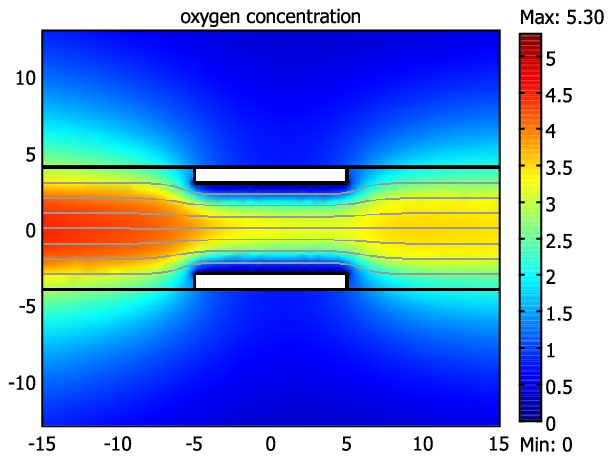}
\end{tabular}
\end{center}
\caption{\figlabel{oxygen concentration}Oxygen concentration in the
tissue and plasma within the vessel. Each diagram shows a cross
section through the vessel and surrounding tissue of length
$30\,\micron$. Typically, this length of vessel contains about 4
cells. The left plots are for the vessel without robots. The right
plots include the 10-micron ringset with pumps, which occupies the
circumferential volume indicated by the white rectangles next to the
vessel wall. The top and bottom plots are for the low and high
demand scenarios of \tbl{scenarios}, respectively. Fluid in the
vessel flows from left to right. Distances along the sides of each
plot are indicated in microns and concentrations on the color bars
are in units of $10^{22}\,\molecule/\meter^3$. The horizontal black
lines are the vessel walls and the gray curves inside the vessel are
fluid flow streamlines.}
\end{figure}

\fig{oxygen concentration} shows the distribution of oxygen in the
tissue and plasma in the vessel near the robots. The robots reduce
the local oxygen concentration far more than the surrounding tissue,
as seen by comparing with the vessel without robots. Most of the
extra oxygen used by the robots comes from the passing blood cells,
which have about 100 times the oxygen concentration of the plasma.
Within the vessel with the robots, the concentration in the plasma
is lowest in the fluid next to the robots. Downstream of the robots
is a recovery region where the concentration increases a bit as
cells respond to the abruptly lowered concentration near the robots.
In the low demand scenario, the concentration in the vessel just
downstream of the robots is somewhat lower than in the surrounding
tissue. Thus in this region, the net movement of oxygen is from the
tissue into the vessel, where the fluid motion transports the oxygen
somewhat downstream before it diffuses back into the tissue. In
effect, part of the oxygen entering the vessel travels through the
tissue around robots to the downstream section of the vessel, in
contrast to the pattern without robots where oxygen is always moving
from the vessel into the surrounding tissue. The streamlines in
\fig{oxygen concentration} show that the laminar flow speeds up as
the fluid passes through the narrower vessel section where the
robots are stationed.

\begin{figure}[th]
\begin{center}
\includegraphics[width=\figwidthS]{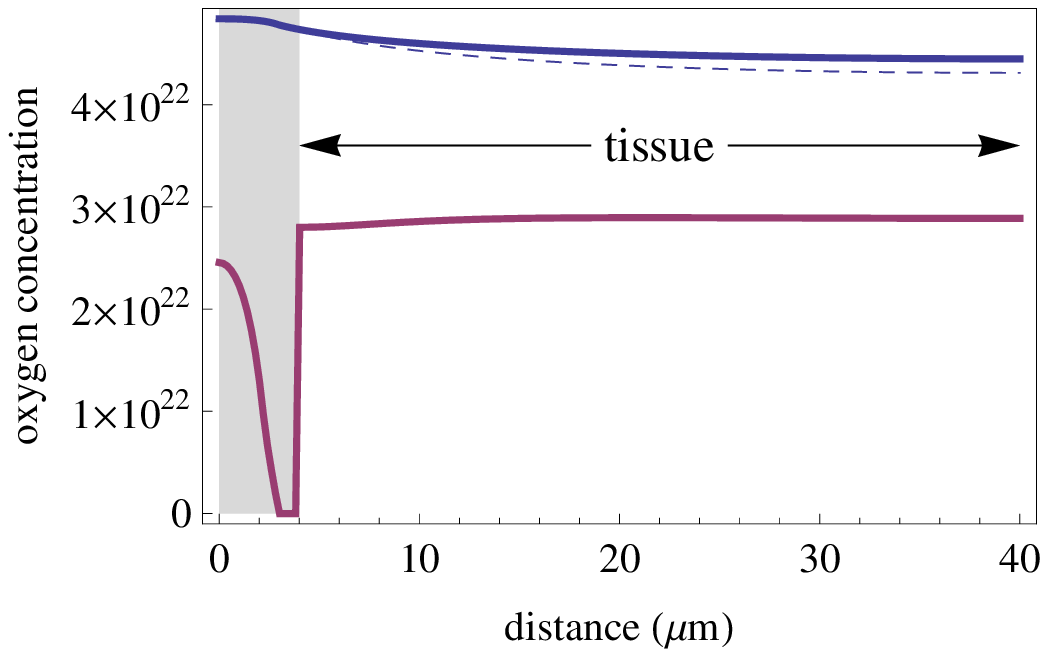} 
\includegraphics[width=\figwidthS]{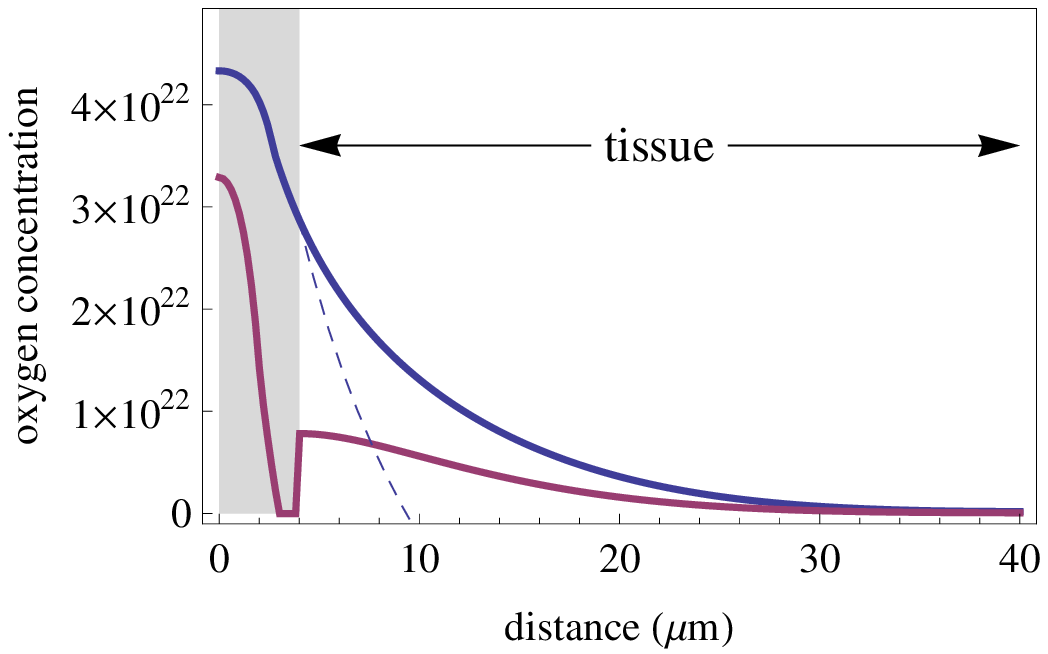}
\end{center}
\caption{\figlabel{cO2 section}Oxygen concentration, in
$\molecule/\meter^3$, along a radial cross section from the center
of the vessel to the outer edge of the tissue region. The cross
section is in the middle of the modeled section of vessel and
tissue, corresponding to a vertical line in the center of each plot
of \fig{oxygen concentration}. The gray area indicates the interior
of the vessel. The left and right plots correspond to the low and
high demand scenarios of \tbl{scenarios}. In each plot, the upper
curve is for the vessel without robots and the lower curve is for
the vessel containing the $10\,\micron$ ringset with pumps. For
comparison, the dashed curves are solutions to the Krogh
model~\cite{krogh19} corresponding to the vessel without robots.}
\end{figure}

\fig{cO2 section} gives another view of how the robots affect the
oxygen concentration in the surrounding tissue. The concentration is
zero at the robot surface facing into the vessel. The robots
decrease the oxygen concentration somewhat but do not affect tissue
power generation much since the concentration remains well above the
threshold where power generation drops significantly, i.e.,
$\ChalfReactionTissue$ given in \tbl{parameters}. However, at large
distances from the vessel in the high demand scenario oxygen
concentration is low enough to significantly decrease tissue power
production. This low level of oxygen also occurs when there are no
robots.

\fig{cO2 section} includes comparison with the simpler Krogh model
of oxygen transport to tissue from vessels without
robots~\cite{krogh19}. The Krogh model assumes constant power
density in the tissue and no diffusion along the vessel direction in
the tissue. For the low demand scenario, the Krogh model results are
close to those from our model. However in the high demand case, the
Krogh model has oxygen concentration drop to zero about
$10\,\micron$ from the vessel, due to the unrealistic assumption of
constant power use rather than the decrease in power use at low
concentrations given by \eq{tissue reaction rate}.

Oxygen flux to the robots ranges from about $10^{19}$ to
$10^{20}\,\molecule/\meter^2/\second$ with the zero-concentration
boundary condition.
Estimates of pump capabilities are up to
$10^{22}\,\molecule/\meter^2/\second$~\cite{freitas99}, which is
more than 100 times the actual flux to the robots. Such pumps could
thereby maintain the zero concentration boundary condition. At an
energy use of $10^{-20}\,\Joule/\molecule$~\cite{freitas99}, the
pumps would require about $1\,\picoWatt$ per robot to handle the
incoming flux, slightly reducing the power benefit of the pumps.
However, much of this pumping energy may be recoverable by adding a
generator using the subsequent expansion of the reaction products to
their lower partial pressure outside the robot~\cite{freitas98}.

\subsection{Robot Power}

This section describes the steady-state power available to the
robots according to our model in various scenarios. We first discuss
the average per robot power in the aggregate, for both high and low
capacity cases, which also indicates the total power available to
the aggregate as a whole. We then show how the power is distributed
among the robots, based on their location in the ringset. Finally,
we illustrate the qualitative features of these results in a
simpler, analytically-solvable model to identify key scaling
relationships between robot design choices and power availability.

\subsubsection{Average Robot Power}

\begin{table}
\begin{center}
\begin{tabular}{l|cc|cc|cc|cc|cc|cc}
robot power generation capacity                  &
\multicolumn{8}{c|}{high capacity}
& \multicolumn{4}{c}{low capacity} \\
inlet concentration $\Coxygen$ {\small ($10^{22}\,/\meter^3$)} \rule{0pt}{10pt} & \multicolumn{4}{c|}{3} & \multicolumn{4}{c|}{7}  & \multicolumn{4}{c}{7}\\
pressure gradient $\pressureGradient$ {\small ($10^5\,\Pascal/\meter$)}\rule{0pt}{10pt} & \multicolumn{2}{c|}{1} & \multicolumn{2}{c|}{5} & \multicolumn{2}{c|}{1} & \multicolumn{2}{c|}{5} & \multicolumn{2}{c|}{1} & \multicolumn{2}{c}{5}  \\
tissue power demand $\PowerTissueMax$
{\small($\kiloWatt/\meter^3$)}\rule{0pt}{10pt} & 4 & 60 &
4 & 60 & 4 & 60 & 4 & 60 & 4 & 60 & 4 & 60\\
\hline 10-micron ringset (with pumps) & 12 &  8 & 14 & 12 & \textbf{17} & 11 & 24 & \textbf{18} & \textbf{17} & 11 & 24 & \textbf{18} \\
10-micron ringset (free diffusion)    & 11 &  7 & 12 & 10 & \textbf{15} & 10 & 22 & \textbf{16} & \textbf{6}  & 3  & 8  & \textbf{6} \\
1-micron ring (with pumps)         & 44 & 27 & 49 & 36 & \textbf{69} & 36 & 99 & \textbf{58} & \textbf{69} & 36 & 99 & \textbf{58} \\
1-micron ring (free diffusion)     & 31 & 19 & 34 & 25 & \textbf{49} & 25 & 71 & \textbf{38} & \textbf{9}  & 4  & 12  & \textbf{7} \\
\end{tabular}
\end{center}
\caption{\tbllabel{behavior: power}Average per-robot power
generation (in picowatts) in various scenarios. Free diffusion is
the ``no pumps'' case. The values in bold correspond to the low and
high demand scenarios of \tbl{scenarios}.}
\end{table}

\tbl{behavior: power} gives the average power generated using the
available oxygen, per robot within the aggregate. As expected,
robots receive more oxygen and hence can generate more power when
inlet concentration is high, fluid speed is high or tissue power
demand is low. In the first two cases, the flow brings oxygen
through the vessel more quickly; in the last case, surrounding
tissue removes less oxygen.
The less than 2-fold decrease in robot power generation in the face
of a larger 2.5-fold decrease in \Oxygen\ inlet concentration from
the arterial to the venous end of the capillary shows that robots
extract more oxygen from red cells than these cells would normally
release while passing the length of the vessel. Thus robots get some
of their oxygen as ``new oxygen'' rather than just taking it from
what the tissues would normally get. This is possible because in
this case robots create steeper concentration gradients than the
tissue does.

The 10-micron ringset with pumps produces about the same power in
the low and high demand scenarios, consuming oxygen at $5\times
10^9\,\molecule/\second$.

Comparing the different aggregate sizes shows lower power
generation, per robot, in the large aggregate compared to the small
one. This arises from the competition among nearby robots for the
oxygen. Nevertheless the larger aggregate, with ten times as many
robots, generates several times as much power in aggregate as the
smaller one. This difference identifies a design choice for
aggregation: larger aggregates have more total power available but
less on a per robot basis.

Robots using pumps generate only modestly more power than robots
relying on diffusion alone in our high capacity design example
(\sect{robot power}). In this case, for robots without pumps, the
power generation site density, $\reactionSiteDensity$, is
sufficiently large that oxygen molecules diffusing into the robot
are mostly consumed by the power generators near the surface of the
robot before they have a chance to diffuse back out of the robot.
For such robots, power generators far from the plasma-facing surface
receive very little oxygen and hence do not add significantly to the
robot power production.

Pumps give higher benefit for isolated rings of robots than for
tightly clustered aggregates. Although not evaluated in the axially
symmetric model used here, pumps may be even more significant for a
single isolated robot on the vessel wall. Such a robot would not be
competing with any other robots for the available oxygen, though
would still compete with nearby tissue.

\subsubsection{Low Capacity Robots}

The low capacity robots have only $1/50^{th}$ the maximum power
generating capability of the high capacity robots discussed above.
Nevertheless, each robot's maximum power is several times larger
than the limit due to available oxygen.
Thus pumps allow the robots to produce the same power as given in
\tbl{behavior: power} for the high capacity robots. The pumps ensure
the absorbed oxygen is completely used by the smaller number of
reaction sites by increasing the concentration of oxygen within the
robots so \eq{reaction rate} gives the same power generation in
spite of the smaller value of $\reactionSiteDensity$.

On the other hand, the smaller number of reaction sites is a
significant limitation for robots without pumps. Comparing with
\tbl{behavior: power} shows pumps improve the average power by
factors of about 3 and 8 for the 10 and 1-micron ringsets,
respectively. Comparing with high capacity robots without pumps
shows the factor of 50 reduction in reaction sites only reduces
average power by factors of about 3 and 5 for the 10 and 1-micron
ringsets, respectively. Thus the reaction sites in the low capacity
scenario are used more effectively than in the high capacity robots:
with a smaller number of sites, each site does not compete as much
with nearby sites for the available oxygen.

Much of the power in robots without pumps is generated near the
plasma-facing surface, where oxygen concentration is largest. In our
case, the power for the high capacity robots is generated primarily
within $100\,\nanometer$ of the robot surface. This observation
suggests that a design with power generating sites placed near this
surface instead of uniformly throughout the robot volume, as we have
assumed, could significantly improve power generation for robots
without pumps. For example, placing all the reaction sites uniformly
within the $1/50^{th}$ of the robot volume nearest the surface would
increase the local reaction site density in that volume by a factor
of 50. For low capacity robots, this placement would increase
$\reactionSiteDensity$ to the same value as the high capacity case,
but only in a narrow volume, within $23\,\nanometer$ of the surface
with the robot geometry we use. Elsewhere in the robot with this
design $\reactionSiteDensity=0$. While we might expect this
concentration to increase power significantly, in fact we find only
a small increase (e.g., 12\% for the 10-micron ringset in the low
demand scenario). Thus concentrating the reaction sites near the
plasma-facing surface does not offer much of a performance
advantage.

\subsubsection{Distribution of Power Among Robots}\sectlabel{power:
distribution}

\begin{figure}[t]
\begin{center}
\includegraphics[width=\figwidthS]{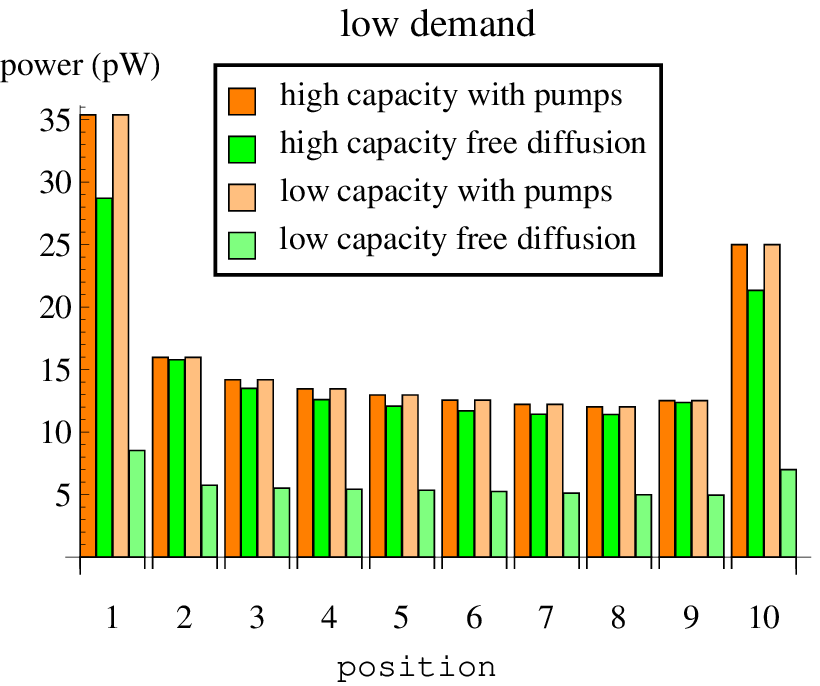}
\includegraphics[width=\figwidthS]{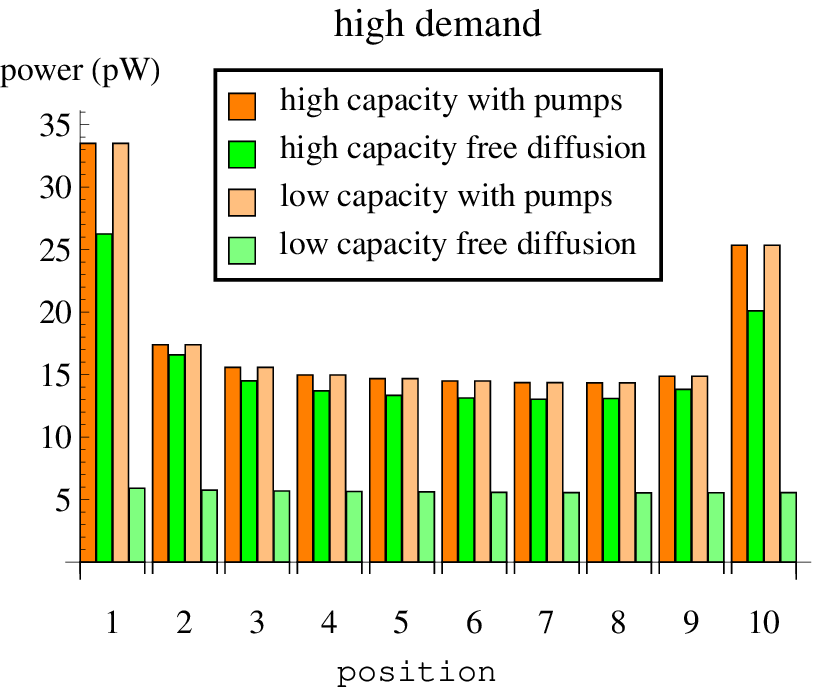}
\end{center}
\caption{\figlabel{power generation}Steady-state power generation,
in picowatts, for robots as a function of their position along the
vessel wall, starting from those at the upstream end of the ringset
(position 1) and continuing to those at the downstream end (position
10). The charts, corresponding to the low and high demand scenarios
of \tbl{scenarios}, compare robots with pumps, absorbing all oxygen
reaching them, with robots relying on free diffusion (i.e., without
pumps), and high and low capacity robots. For robots with pumps,
power for high and low capacity cases are the same.}
\end{figure}

While all robots in a single ring have the same power due to the
assumption of axial symmetry, \fig{power generation} shows that
power varies with ring position in the 10-micron ringset. The robots
at the upstream edge of the aggregate receive more oxygen than the
other robots and hence produce more power. Power generation does not
decrease monotonically along the vessel: robots at the downstream
edge have somewhat more available oxygen than those in the middle of
the aggregate since robots at the edge of the aggregate have less
competition for oxygen. \fig{power generation} shows significantly
larger benefits of pumps at the edges of multi-ring aggregates than
in their middle sections, especially in the low demand scenario.

In the scenarios described above, robots produce power from all the
available oxygen. This is appropriate for applications requiring as
much power as possible for the aggregate as a whole. At the other
extreme, an application requiring the same behavior from all robots
in the aggregate would be limited by the robots with the least
available power. This would be the case for identical robots, all of
which perform the same task and hence use the same power. In this
case, the robots could increase performance by transferring power
from those at the edges of the aggregate to those in the middle.
Such transfer could take place after generation, e.g., via shared
electric current, or prior to generation by transfer of oxygen among
neighboring robots. However, such internal transfer would require
additional hardware capabilities. For robots with pumps, an
alternative transfer method is for robots near the edge of the
aggregate to run their pumps at lower capacity and thus avoid
collecting all the oxygen arriving at their surfaces. This
uncollected oxygen would then be available for other robots, though
some of this oxygen would be transported by the fluid past the
robots or be captured by the tissue rather than other robots. This
approach increases power to robots in the middle of the aggregate
without requiring additional hardware for internal transfers between
robots, but at the cost of somewhat lower total power for the
aggregate. Increasing as much as possible the power to the robots
with the least power leads to a uniform distribution of power among
the robots.

To quantify the trade-off between total power and its uniformity
among the robots, we consider all robots setting each of the pumps
on their surfaces to operate at the same rate and the pumps
uniformly distributed over the surface. This gives a uniform flux of
oxygen over the entire surface of all the robots. The largest
possible value for this uniform flux, and hence the largest power
for the aggregate, occurs when the minimum oxygen concentration on
the robot surfaces is zero -- at that point, the robot whose surface
includes the location of zero concentration cannot further increase
its uniform flux.
For example, in the low demand scenario the maximum value for this
uniform flux is approximately $2.22 \times
10^{19}\,\molecule/\meter^2/\second$. Compared to the situation in
\fig{power generation}, this uniform flux gives significantly lower
power (39\% leading ring, 56\% trailing ring) for the robots at the
edges of the aggregate, somewhat lower power for robots in positions
2 and 3 (87\% and 98\%, respectively) and somewhat more power
(ranging from 104\% to 116\%) for the other robots. The combination
of these changes gives a total of $84\%$ of the power for the
aggregate when every robot collects all the oxygen reaching its
surface. The minimum power per robot increases from $12\,\picoWatt$
to $14\,\picoWatt$.
Robots could slightly increase power by accepting some nonuniformity
of flux over each surface while maintaining the same total flux to
each robot. This would occur when the minimum oxygen concentration
on the entire length of the robot surfaces in a particular robot
ring is zero.

Using this approach to uniform power in practice would require the
robots to determine the maximum rate they can operate their pumps
while achieving uniform power distribution. This rate would vary
with tissue demand, and also over time as cells pass the robots. A
simple control protocol is for each robot to adjust its pump rate up
or down according to whether its power generation is below or above
that of its neighbors, respectively. When concentration reaches zero
on one robot, increasing pump rate at that location would not
increase power generation.
Communicating information for this protocol is likely simpler than
the hardware required to internally transfer power or oxygen among
robots, but also requires each robot is able to measure its power
generation rate. Such measurements and communication would give an
effective control provided they operate rapidly compared to the time
over which oxygen flux changes, e.g., as cells pass the robots on
millisecond time scales. Longer reaction times could lead to
oscillations or chaotic behavior~\cite{hogg04b}.

A second approach to achieving a more uniform distribution of power
is to space the robots at some distance from each other on the
vessel wall. This approach would be suitable if the aggregated
robots do not need physical contact to achieve their task. For
example, somewhat separating the robots would allow a relatively
small number to span a distance along the vessel wall larger than
the size of a single cell passing through the vessel. This aggregate
would always have at least some robots between successive cells.
Communicating sensor readings among the robots would then ensure the
response, e.g., releasing chemicals, is not affected by misleading
sensor values due to the passage of a single cell, giving greater
stability and reliability without the need for delaying response due
to averaging over sensor readings as an alternative approach to
accounting for passing cells. Another example for spaced robots is
for directional acoustic communication, at distances of about
$100\,\micron$~\cite{freitas99}. Achieving directional control
requires acoustic sources extending over distances comparable to or
larger than the sound wavelength. Plausible acoustic communication
between nanorobots involves wavelengths of tens of
microns~\cite{freitas99}.

As a quantitative example of the benefit of spacing robots in the
context of our axially symmetric model, we consider a set of rings
of robots spaced apart along the vessel wall. When the distance
between successive rings is sufficiently large, the power for each
ring would be close to that of the isolated 1-micron ring given in
\tbl{behavior: power}. For example, the power for the low demand
scenario in the 1-micron ring of high-capacity robots decreases from
$69\,\picoWatt$ at high inlet concentration to $44\,\picoWatt$ at
low inlet concentration, which spans the range of power for a modest
number of widely spaced 1-micron rings within a single vessel. As
described in \sect{tissue power and heating}, oxygen absorption by
robots can affect concentration over a few tens of microns upstream
of those robots. Thus separating robot rings by, say, $100\,\micron$
will give power close to that of the isolated rings, with a gradual
decrease in power for successive rings due to the decreasing cell
saturation along the vessel.

A third approach to reducing variation in robot power, on average,
is through changing pump rates in time. For example, adjacent
nanorobot rings could operate with counterphased 50\% duty cycles,
with one ring and its second nearest neighbor ring using pumps while
the intervening nearest neighbor has its pumps off and does not
absorb oxygen. The alternating rings of robots would switch pumps on
and off.
In this case, robots would have larger power than seen in \fig{power
generation} for the half of the time they are active, and zero power
for the other half. This temporal approach would not be suitable for
tasks requiring all robots to have the same power
\emph{simultaneously}, but would be useful for tasks requiring
higher burst power from robots throughout the aggregate where the
robots are unable to store oxygen or power for later use. Provided
the duty cycle is sufficiently long, our steady-state model can
quantify the resulting power distribution.
For example, in the low demand scenario, total flux for the
aggregate is 79\% of that when every robot collects all the oxygen
reaching its surface, and the minimum power per robot \emph{drops}
from $12\,\picoWatt$ to $10\,\picoWatt$. Thus, when averaged over
the duty cycle, this temporal technique reduces total power without
benefiting the robots receiving the minimum power.
In this case, the temporal approach does not improve minimum robot
power (on average) since the power gain to a robot while its
neighbors are off is less than a factor of 2, which does not
compensate for the loss due to each robot being off for half the
time.
Applying the steady-state model to this temporal variation in robot
activity requires the duty cycle be long enough for the system to
reach steady-state behavior after each switch between the active
subset of robots, and that the switching time is short compared to
the duty cycle so most of the robots' power arises during the
steady-state portions of the cycle between switching. Diffusion
provides one lower bound on this time: when neighboring robots
switch pumps from on to off or vice versa, the characteristic
diffusion time for oxygen over the distance between next nearest
neighbors (one micron) is about $0.1\,\millisecond$.
Adjustments in cell saturation for the 1-micron shift in the
location of the active robots between each half of the duty cycle is
a further limitation on the duty cycle time for the validity of the
steady-state model, though this is likely to be minimal since the
cells are separated from the robots by the plasma gap in the fluid.
Since the steady-state model averages over the position of passing
cells, another lower bound on the duty cycle arises from the time
for a cell to pass the robots. From the speeds in \tbl{parameters},
this time is at least $100\,\millisecond$.

\subsubsection{Analytical Model for an Isolated Spherical Robot}

The dependence of robot power on design parameters described above
may appear contrary to simple intuitions. First, one might expect
that the 10-micron ringset, with ten times the surface area in
contact with the plasma, would absorb about ten times as much oxygen
as the 1-micron ring. Instead we find only about a factor of 2 to 4
increase. Second, the benefit of pumps, less than a factor of 2 for
the high capacity robots, may seem surprisingly small. Third, the
low capacity robots, with $1/50^{th}$ the reaction sites of the high
capacity robots, nevertheless generate about $1/5^{th}$ as much
power as high capacity robots in the case with no pumps. And
finally, in spite of the higher concentration near the robot surface
than deep inside the robot when there are no pumps, increasing the
reaction site density by placing all the reaction sites near the
robot surface gives little benefit. While the specific values of
these designs depend on the geometry and environment used in our
model, these general features of small robots obtaining power
through diffusion apply in other situations as well.

In this section we illustrate how these consequences of design
choices arise in the context of a scenario for which the diffusion
equation has a simple analytic solution, thereby identifying key
physical effects leading to these behaviors. Specifically, we
consider an isolated spherical robot of radius $a$ in a stationary
fluid with oxygen concentration $C$ far from the sphere.

Such a sphere with a fully absorbing surface collects oxygen at a
rate $4 \pi \Doxygen a C$~\cite{berg93}. This expression illustrates
a key property of diffusive capture: the rate depends not on the
object's surface area but on its size. This behavior, which also
applies to other shapes~\cite{berg93}, arises because while larger
objects have greater surface areas they also encounter smaller
concentration gradients.
As a quantitative example, taking the sphere to have the same volume
as the robot, i.e., $(4/3)\pi a^3 = \Vrobot$ given in \tbl{robot
parameters}, the oxygen absorbed by the sphere generates $320$ and
$750\,\picoWatt$ for $C$ equal to the low and high inlet oxygen
concentrations in the plasma from \tbl{parameters}, respectively.
These power values are larger than for robots on the vessel wall
described above. Unlike the sphere in a stationary fluid, the
aggregated robots compete with each other for the oxygen, the fluid
moves some of the oxygen past the robots before they have a chance
to absorb it, and the surrounding tissue also consumes some of the
oxygen. The replenishment of oxygen from the passing blood cells is
not sufficient to counterbalance these effects.

The spherical robot also indicates the benefit of pumps. The fully
absorbing sphere, with a zero concentration boundary condition at
the surface, corresponds to using pumps. For robots without pumps,
an approximation to \eq{reaction rate} allows a simple solution.
Specifically, since the Michaelis-Menten constant for the robot
power generators, $\ChalfReaction$, is much larger than the oxygen
concentrations (e.g., as seen in \fig{oxygen concentration}), robot
power generation from \eq{reaction rate} is approximately
$\PowerRobot \approx (\reactionEnergy \reactionSiteDensity
\reactionRate/\ChalfReaction)\Coxygen$. Dividing by
$\reactionEnergy/6$ gives the oxygen consumption rate density as
$\gamma \Coxygen$ where $\gamma = 6 \reactionSiteDensity
\reactionRate/\ChalfReaction$. Solving the diffusion equation,
\eq{diffusion}, for a sphere in a stationary fluid with
concentration $C$ far from the sphere, with free diffusion through
the sphere's surface and reaction rate density $\gamma \Coxygen$
inside the sphere, gives the rate oxygen is absorbed by the sphere
(and hence reacted to produce power) as~\cite{hogg09} $ 4 \pi
\Doxygen C (a - \mu \tanh(a/\mu))$ where $\mu =
\sqrt{\Doxygen/\gamma} $. Thus free diffusion produces a fraction
\begin{equation}\eqlabel{f mu}
f_{\mu} = 1 - \frac{\mu}{a} \tanh\left(\frac{a}{\mu}\right)
\end{equation}
of the power produced by the fully absorbing sphere. The distance
$\mu$ is roughly the average distance an oxygen molecule diffuses in
the time a power generation site consumes an oxygen molecule. When a
freely diffusing molecule inside the robot has a high chance to
diffuse out of the robot before it reacts ($\mu$ large compared to
$a$), $f_{\mu}$ is small so pumps provide a significant increase in
power. Conversely, when $\mu$ is small compared to $a$, pumps
provide little benefit: the large number of reaction sites ensure
the robot consumes almost all the diffusing oxygen reaching its
surface.

This argument illustrates a tradeoff between using pumps to keep
oxygen within the robot and the number of power generators. In
particular, if internal reaction sites are easy to implement, then
robots with many reaction sites and no pumps would be a reasonable
design choice. Conversely, if reaction sites are difficult to
implement while pumps are easy, then robots with pumps and few
reaction sites would be a better choice.

A caveat for robots with few power generating sites is that \eq{f
mu} applies when oxygen consumption is linear in the concentration,
as given by $\gamma \Coxygen$. This expression allows arbitrarily
increasing the reaction rate by increasing the concentration, no
matter how small the number of reaction sites. This linearity is a
good approximation of \eq{reaction rate} only when $\Coxygen \ll
\ChalfReaction$. At larger concentrations the power density
saturates at $\reactionEnergy \reactionSiteDensity \reactionRate$.
When $\reactionSiteDensity$ is sufficiently small, this limit is
below the power that could be produced from all the oxygen that a
fully absorbing sphere collects. Thus, in practice, the benefit of
using pumps estimated from the linear reaction rate, $G =
f_{\mu}^{-1}$, is limited by this bound when $\reactionSiteDensity$
is small.

As an example, for a spherical robot with the high capacity reaction
site density of \tbl{robot parameters}, $\gamma = 1.8\times
10^4/\second$ and $\mu = 0.33\,\micron$, with the fairly modest
benefit of pumps $G = 2.0$. The low capacity robots have $\gamma =
360/\second$ and $\mu = 2.4\,\micron$ with $G=42$. In this case, the
limit due to the maximum reaction rate of \eq{reaction rate}
applies, somewhat limiting the benefit of pumps to a factor of $34$,
but pumps still offer considerable benefit. These values for the
benefits of pumps are somewhat larger than seen with our model for
robots on the vessel wall. Nevertheless, the spherical example
identifies the key physical properties influencing power generation
with and without pumps, and how they vary with robot design choices.

\eq{f mu} also illustrates why power in the low capacity robots is
not as small as one might expect based on the reduction in reaction
sites by a factor of 50. While the value of $\gamma$ is proportional
to \reactionSiteDensity, the typical diffusion distance $\mu$ varies
as $1/\sqrt{\reactionSiteDensity}$, so a decrease in reaction sites
by a factor of 50 only increases $\mu$ by about a factor of 7. The
square root dependence arises from the fundamental property of
diffusion: typical distance a diffusing particle travels grows only
with the square root of the time. The modest change in diffusion
distance, combined with \eq{f mu}, gives a smaller decrease in power
than the factor of 50 decrease in capacity. The low capacity robot
has higher concentrations throughout the sphere, so each reaction
site operates more rapidly than in the high capacity case. This
increase partially offsets the decrease in the number of reaction
sites.

Without pumps, the higher oxygen concentration near the sphere's
surface than near its center means much of the power generation
takes place close to the surface. Thus we can expect an increase in
power by placing reaction sites close to the surface rather than
uniformly distributed throughout the sphere. Consistent with the
results from the model described in \sect{model}, evaluating
\eq{diffusion} with the reaction confined to a spherical shell shows
only a modest benefit compared to a uniform distribution. The
benefit is larger for a thinner shell and is determined by the same
ratio, $a/\mu$, appearing in \eq{f mu}. In particular, the largest
benefit of using a thin shell, only $12\%$, occurs for $a/\mu
\approx 3.5$. The parameters for the high and low capacity robots
are somewhat below this optimal value, giving only $10\%$ and less
than $1\%$ benefit from a thin shell, respectively, for the sphere.
These modest improvements correspond to the small benefits of using
a thin shell seen in the solution to our model for both high and low
capacity robots. Hence the solution of \eq{diffusion} for the sphere
illustrates how, with a fixed number of reaction sites,
concentrating them near the robot surface provides only limited
benefit. The benefit of the higher reaction site density in the
shell is almost entirely offset by the shorter distance molecules
need to diffuse to escape from the thin reactive region. That is,
the benefit of placing all the reaction sites in a thin shell arises
from two competing effects. When $\mu$ is large (low capacity) the
concentration is only slightly higher near the surface than well
inside the sphere. So there is little benefit from placing the
reaction sites closer to the surface. On the other hand, when $\mu$
is small (high capacity), even uniformly distributed reaction sites
manage to consume most of the arriving oxygen, giving near-zero
concentration at the surface of the sphere and little scope for
further improvement by concentrating the reaction sites. Thus the
largest, though still modest, benefit for a shell design is for
intermediate values of $a/\mu$.

\subsection{Oxygen Replenishment from Passing Cells}

\begin{figure}
\begin{center}
\includegraphics[width=\figwidth]{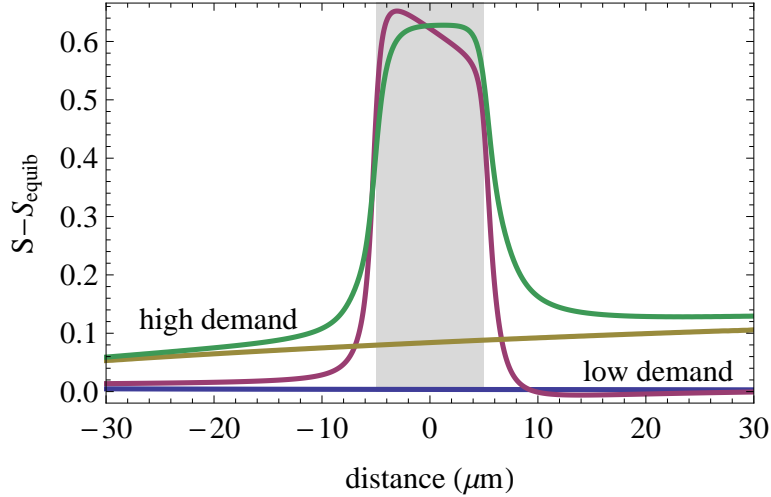}
\end{center}
\caption{\figlabel{S-Sequib}Deviation from equilibrium oxygen
saturation in cells, $S-\Sequib$, along the boundary between the
cell and cell-free portions of the fluid illustrated in
\fig{geometry}, as a function of distance along that boundary. A
deviation of zero indicates the oxygen held in the cells is in
perfect equilibrium with the surrounding plasma. Saturation ranges
between 0 and 1. The curves correspond to the low and high demand
scenarios of \tbl{scenarios}, when robots are present (upper curves)
or absent (lower curves). For the vessel without robots and low
demand, $S-\Sequib$ is indistinguishable from zero on the scale of
the plot. The gray band indicates the 10-micron length of the vessel
wall in which the robots are stationed, within the 60-micron length
of the capillary illustrated.}
\end{figure}

The high power density of the robots creates a steep gradient of
oxygen concentration in the plasma. Thus, unlike the minor role for
nonequilibrium oxygen release in tissue~\cite{popel89}, the small
size of the robots makes passing red cells vary significantly from
equilibrium with the concentration in the plasma. \fig{S-Sequib}
illustrates this behavior, using one measure of the amount of
disequilibration: the difference between saturation $S$ and the
equilibrium value $\Sequib$ corresponding to the local concentration
of oxygen in the plasma, as given by \eq{equilibrium saturation}. We
compare with a vessel without robots, in which the blood cells
remain close to equilibrium.

\fig{S-Sequib} shows that the kinetics of oxygen release from red
cells plays an important role in limiting the oxygen available to
the robots. However, the region of significant disequilibration is
fairly small, extending only a few microns from the robots.

\subsection{Tissue Power and Heating}\sectlabel{tissue power and
heating}

\begin{figure}[t]
\begin{center}
\includegraphics[width=\figwidth]{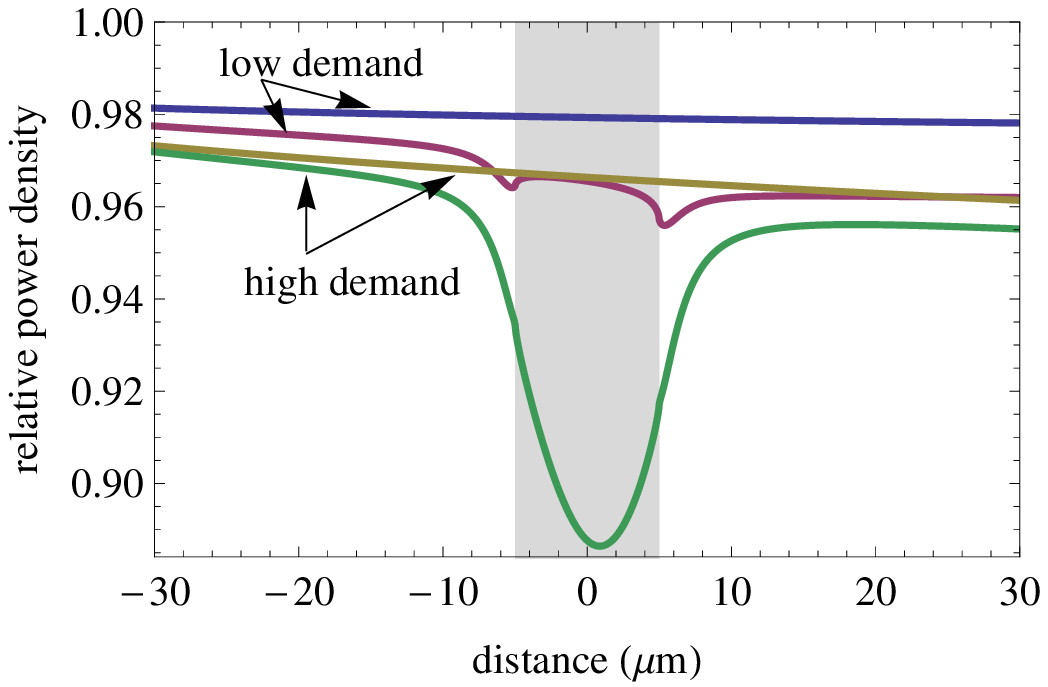}
\end{center}
\caption{\figlabel{tissue power}Power density in tissue next to the
vessel wall relative to maximum demand, i.e., the ratio
$\PowerTissue/\PowerTissueMax$ from \eq{tissue reaction rate}, as a
function of position along the vessel. The curves are for the low
and high demand scenarios of \tbl{scenarios}. The two lines at the
top are for the vessel without robots and the lower curves are for
the 10-micron ringsets. In each case, the curve with higher values
corresponds to the low power demand scenario. The gray band
indicates the 10-micron length of the vessel wall in which the
robots are stationed, within the 60-micron length of the capillary
illustrated.}
\end{figure}

\begin{figure}
\begin{center}
\includegraphics[width=\figwidth]{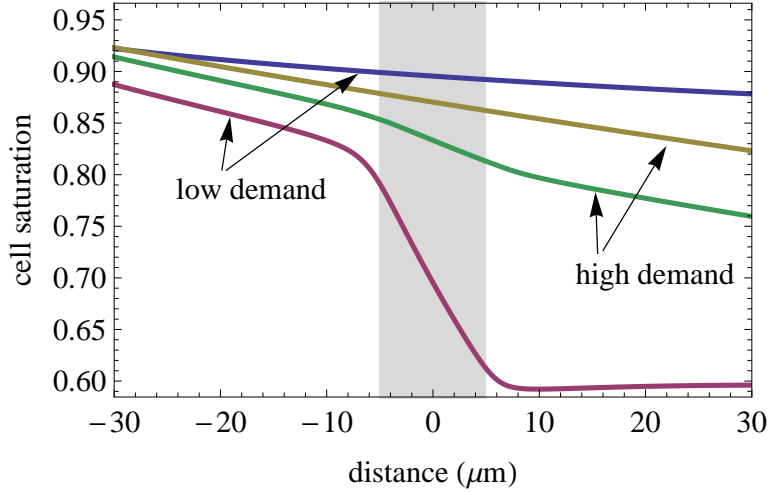}
\end{center}
\caption{\figlabel{saturation}Cell saturation $S$ as a function of
distance along the vessel. Saturation ranges between 0 and 1. The
curves correspond to the low and high demand scenarios of
\tbl{scenarios}, with the upper curves of each pair corresponding to
a vessel without robots. The gray band indicates the 10-micron
length of the vessel wall in which the robots are stationed, within
the 60-micron length of the capillary illustrated.}
\end{figure}

The robots affect tissue power in two ways. First, the robots
compete for oxygen with nearby tissue. Second, the robots consume
oxygen from passing blood cells, thereby leaving less for tissue
downstream of the robots.

For the effect on nearby tissue, \fig{tissue power} shows how tissue
power density varies next to the vessel wall. In the vessel without
robots, power density declines slightly with distance along the
vessel as the tissue consumes oxygen from the blood. The total
reduction in tissue power density is fairly modest, less than 10\%
even for high power demand in the tissues. The relative reduction is
less for tissue at larger distances from the vessel, though such
tissue has lower power generation due to less oxygen reaching tissue
far from the vessel. This reduction arises both from direct
competition by the robots for available oxygen and the physical
blockage of the capillary wall, forcing surrounding tissue to rely
on oxygen diffusing a longer distance from unblocked sections of the
wall. In the low demand case, direct competition is the major
factor, as seen by the dips in the power density at each end of the
aggregate, where the absorbing flux is highest. In the high demand
case, the tissue's consumption reduces the amount of oxygen
diffusing through the tissue on either side of the aggregate, giving
the larger drop in tissue power density in the middle of the
aggregate.

For longer range consequences, \fig{saturation} shows how the oxygen
saturation in the blood cells changes as they pass the robots.
Slowly moving cells (in the low demand scenario) are substantially
depleted while passing the robots, even though tissue power demand
in this scenario is low. This depletion arises from the cells
remaining near the robots a relatively long time as cells move
slowly with the fluid. The resulting saturation shown in the figure,
around $0.6$, is below the equilibrium saturation ($S=0.7$) for
typical concentrations at the venous end of capillaries, given in
\tbl{parameters}. Thus in the low demand scenario, the robots remove
more oxygen from passing cells than occurs during their full transit
of a vessel containing no robots. In this scenario, the tissue has
low power demand, so the depletion of oxygen from the cells may have
limited effect on tissue along the vessel downstream of the robots.
However, this reduction could significantly limit the number of
robots that can be simultaneously present inside a given capillary.

Another observation from \fig{saturation} is a significant decrease
in cell saturation a short distance \emph{upstream} of the robots in
the low demand scenario. We can understand this behavior in terms of
the Peclet number, which characterizes the relative importance of
convection and diffusion over various distances~\cite{squires05}. In
particular, $D/\vAvg$ is the distance at which diffusion and
convection have about the same effect on mass transport in a moving
fluid. At significantly longer distances, convection is the dominant
effect and absorption of oxygen at a given location in the vessel
has little effect on upstream concentrations over such distances. In
our scenarios, $D/\vAvg$ ranges from $\sim 2\,\micron$ (high demand)
to $\sim 10\,\micron$ (low demand). Thus the oxygen concentration in
the plasma is significantly affected by the robots over a few tens
of microns upstream of their location. Cell saturation remains close
to equilibrium in this upstream region (\fig{S-Sequib}), hence the
reduced oxygen concentration in the plasma lowers cell saturation in
this region upstream of the robots (\fig{saturation}). This distance
is also relevant for spacing rings of robots far enough apart to
achieve nearly uniform power, as described in \sect{power:
distribution}.

The devices in this example have a volume of $\Vrobot \approx
1\,\micron^3$ so the robot power generation corresponds to power
densities around $10^7 \,\Watt/\meter^3$, several orders of
magnitude larger than power densities in tissue, raising concerns of
possible significant tissue heating by the robots. However, for the
isolated aggregate used in this scenario, waste heat due to the
robots' power generation is rapidly removed, resulting in negligible
maximum temperature elevation of about
$10^{-4}\,^{\circ}\mathrm{C}$.

\subsection{Fluid Flow and Forces on the Robots}

The robots change the fluid flow by constricting the vessel. With
the same pressure difference as a vessel without robots, as used in
our model, this constriction results in somewhat lower flow speed
through the vessel. Specifically, the one and ten-micron long
aggregates reduce flow speed by 6\% and 20\%, respectively.

The fluid moving past the robots exerts a force on them. To remain
on the wall the robots must resist this force through their
attachment to the vessel wall~\cite{freitas99}. This force is a
combination of pressure difference, between the upstream and
downstream ends of the aggregate, and viscous drag. For the laminar
flow the force $f$ is linear in the pressure gradient
$\pressureGradient$ imposed on the vessel: $f = a \pressureGradient$
where $a = 1.56\times 10^{-15}\,\meter^3$ and $5.02\times
10^{-16}\,\meter^3$ for the $10\,\micron$ and $1\,\micron$ ringsets,
respectively. For example, the flow imposes a force of
$160\,\picoNewton$ on the $10\,\micron$ ringset when the pressure
gradient is $10^5\,\Pascal/\meter$. The $10\,\micron$ ringset
experiences about three times the force of the $1\,\micron$ ring,
but covers ten times the surface area. Thus the larger aggregate
requires about one-third the attachment force per robot. Applied
forces can affect cells~\cite{discher05}. In particular, endothelial
cells use forces as a trigger for new vessel
growth~\cite{deroanne01}, which is important for modeling changes in
the vessels over longer time scales than we consider in this
paper~\cite{szczerba06}.

\section{Discussion}\sectlabel{discussion}

The scenarios of this paper illustrate how various physical
properties affect robot power generation.
Robots about one micron in size positioned in rings on capillary
walls could generate a few tens of picowatts in steady state from
oxygen and glucose scavenged locally from the bloodstream.
Aggregates can combine their oxygen intake for tasks requiring
higher sustained power generation. The resulting high power
densities do not significantly heat the surrounding tissue, but do
introduce steep gradients in oxygen concentration due to the
relatively slow reaction kinetics of oxygen release from red cells.
The robots reduce oxygen concentration in nearby tissues, but
generally not enough to significantly affect tissue power
generation.

The fraction of the generated power available for useful activity
within the robot depends on the efficiency of the glucose engine
design, with $\sim 50\%$ a reasonable estimate for fuel
cells~\cite{freitas99}. The robots will have
$5\mbox{--}30\,\picoWatt$ of usable steady-state power while on the
vessel wall. As one indication of the usefulness of this power for
computation, current nanoscale electronics and sensors have an
energy cost per logic operation or sensor switching event of a few
hundred $\BoltzmannConstant T$~\cite{wang05}. While future
technology should enable lower energy use~\cite{freitas99}, even
with $500\,\BoltzmannConstant T \approx 2\times 10^{-18}J$ the
available power from circulating oxygen could support several
million computational operations per second. At the size of these
robots, significant movements of blood cells and chemical transport
occur on millisecond time scales. Thus the power could support
thousands of computational operations, e.g., for chemical pattern
recognition, in this time frame. The aggregated robots could share
sensor information and CPU cycles, thereby increasing this
capability by a factor of tens to hundreds.

The robots need not generate power as fast as they receive oxygen,
but could instead store oxygen received over time to enable bursts
of activity as they detect events of interest. Robots with pumps
have a significant advantage in burst-power applications because
pumps enable long-term high-concentration onboard gas accumulation
to support brief periods of near maximal power generation. As an
example, in our scenarios individual robots with pumps receive about
$10^8\,\molecule/\second$. If instead of using this oxygen for
immediate power generation, the robot stored the oxygen received
over one second, it would have enough to run the power generators at
near maximal rate (giving about $10^4\,\picoWatt$) for several
milliseconds. By contrast, robots without pumps would only have a
modest benefit from oxygen diffusing into the robot, achieving a
concentration equal to the ambient concentration in the surrounding
plasma as given in \tbl{parameters}. This concentration could only
support generating several hundred picowatts for about a tenth of a
millisecond. Thus while pumps may give only modest improvement for
steady-state power generation, they can significantly increase power
available in short bursts.

Our model could be extended to estimate the amount of onboard
storage that would be required to avoid pathological conditions
related to \Oxygen\ competition between tissues and nanorobots. In
particular, larger aggregates would deplete oxygen over longer
distances for which diffusion through the tissue from upstream of
the robots would be insufficient. Furthermore, larger aggregates of
tightly-spaced robots would block transport from the capillary into
the surrounding tissue even if the robots did not use much oxygen.
Onboard oxygen storage would allow higher transient power densities
for the robots, though this could lead to heating issues for larger
aggregates.
To estimate the potential for onboard storage, a ringset containing
200 robots with volume $200\Vrobot \approx 220\,\micron^3$ of which
10\% is devoted to compressed \Oxygen\ storage at 1000 atmospheres
at body temperature can store about $5\times 10^{11}$ molecules of
\Oxygen\ in the aggregate. The incoming flow in the capillary
provides about $0.2\mbox{--}1\times 10^{11}$ molecules per second,
depending on the flow speed, of which about $1/4$ to $1/2$ is
available to the tissue and robots. This means the oxygen stored in
the aggregate is equivalent to only several seconds of oxygen
delivery through the vessel. Thus oxygen storage in the robots
themselves can not significantly increase mission duration, though
such storage might be useful for short-term (i.e., a few seconds)
load leveling functions (e.g., maintaining function during temporary
capillary blockage due to white cell passage).

Alternatively, the aggregated robots could have oxygen supplemented
with a modest circulating population of
respirocytes~\cite{freitas98}, i.e., $1\,\micron$ spherical robots
able to carry oxygen to tissues far more effectively than red blood
cells. Such robots would continuously and entirely eliminate any
oxygen depletion regions in the tissue due to robot power
generation, and allow higher robot power generation since oxygen
would no longer be such a limiting factor.
Such machines could not only carry significantly more oxygen than
red blood cells, but would also respond more quickly to abrupt
decreases in partial pressure due to consumption by aggregated
robots on vessel walls. For example, sensors should be able to
detect the drop in concentration of the size we see near the robots
-- e.g., $(3 \mbox{ to } 2) \times 10^{22}\,\molecule/\meter^3$
 (or about $40-20\, \mbox{mmHg}$) -- within a millisecond~\cite{freitas99}.
This time is short enough that the machines will have moved only
about a micron and so will still be near the robots. Once they
detect the pressure drop, the machines could release oxygen rapidly,
up to $1.5\times 10^8 \,\molecule/\second$~\cite{freitas98}, while
passing near the aggregated robots. However, in practice the release
rate is constrained by the effervescence limit in plasma to about
$2.5\times 10^7\, \molecule/\second$~\cite{freitas99}. An
interesting question for future work is evaluating how much of this
released oxygen reaches the robots on the vessel wall, which will
depend on how close to the vessel wall the fluid places the
respirocytes.
In this situation, aggregated robots could also
communicate~\cite{freitas99} to passing respirocytes to activate or
suppress their oxygen delivery, depending on the task at hand. Thus
both the oxygen handling capabilities of respirocytes, giving faster
kinetics than red cells and larger storage capacity, and the
possibility of communication provide examples of the flexibility of
small devices with programmable control. Moreover, this scenario
illustrates the benefits of mixing robots with differing hardware
capabilities.

Our model considers static aggregates on vessel walls, but could be
extended to study power availability for aggregates that move along
the walls~\cite{freitas96}. Another significant scenario is robots
moving passively with the fluid, where they could draw oxygen from
the surrounding plasma. The oxygen unloading model used here could
evaluate how rapidly nearby cells would replenish oxygen in the
plasma as the cells and robots move through the capillary.

We treat environmental parameters (e.g., fluid flow speed and tissue
oxygen demand) as fixed by the surroundings. Beyond the local
changes in the robots' environment described by the model, sustained
use of these robots could induce larger scale responses. For
example, the increased use of oxygen by the robots could lead to
increased blood flow, as occurs with, say, exercising muscles, by
increased pressure to drive the fluid at higher speed or dilation of
the vessels. The local oxygen deficits due to high robot power use
are smaller in scale than higher tissue demand (e.g., from increased
activity in a muscle). Thus an important open question is whether
localized robot oxygen consumption over a long period of time can
initiate a less localized response to increase flow in the vessels.

The possibility of large-scale responses to robot activity raises a
broader issue for nanomedicine treatment design when technology
allows altering the normal correlation among physical quantities
used for signalling in the body~\cite{freitas99,freitas06}. An
example at larger scales is the response to low oxygen mediated by
excess carbon dioxide in the blood, which can lead to edema and
other difficulties for people at high altitudes~\cite{schoene08}.
In terms of downstream consequences of the robots' oxygen use, low
saturation of cells leaving an isolated capillary should not be a
problem because the bulk of oxygen exchange occurs in the capillary
bed, not in the larger collecting vessels.  However, cells reaching
low saturation before exiting the capillary would produce localized
anoxia in the tissue near the end of the capillary. This could be
relieved in part by oxygen diffusion from neighboring tissue cells
if the anoxic region is not too large or too severe. Specific
effects of such localized anoxia remain to be fully identified.
Whole capillaries subjected to ischemic conditions over a period of
days remodel themselves, e.g., by adding new vascular branches and
by increasing the tortuosity of existing vessels~\cite{bailey08}.
This observed behavior is likely to be a localized (i.e.,
cell-level) response, hence we might expect such a response if a
portion of a capillary downstream of the robots was driven into
ischemic conditions. There could also be a localized inflammatory
response to a large enough number of capillary-wall endothelial
cells under stress, especially for cells stressed to the point of
apoptosis, but moderate ischemia alone seems unlikely to generate
this response~\cite{brown03a}.
Various chemicals (e.g., adrenalin) make the heart pump faster and
thus drive the blood at higher speed. Other chemicals (such as NO,
PGD2 and LTD4) dilate the vessels. These chemicals can produce
significant activity in the endothelial cells that line (and thus
form the tube geometry of) the capillary vessel, so their influence
can be fairly direct and quick~\cite{soter83,sekar06}.
Similarly, a large robot population constantly drawing excess oxygen
supply could induce elevated erythropoietin secretion (if
unregulated by the robots), increasing red cell production in the
erythroid marrow~\cite{freitas98,erslev95}.

Direct heating is not a problem with aggregates of the size
considered here, in spite of their high power density compared to
tissue. For the large aggregate we examined (tightly covering
$10\,\micron$ along the vessel wall), oxygen diffusion through the
tissue from regions upstream and downstream of the robots provided
oxygen to the tissue outside the section of the vessel blocked by
the robots. Larger aggregates, especially if tightly packed, would
significantly reduce oxygen in the tissues even if the robots used
little power themselves, simply due to their covering the vessel
wall over a long enough distance that diffusion through the tissue
from unblocked regions is no longer effective. The inducement of
nonlinear tissue thermal responses (e.g., inflammation or fever) due
to the heat generated by larger aggregates or multiple aggregates in
nearby capillaries is an important question for future work.

Nanorobots parked or crawling along the luminal surface of the
vessel may activate mechanosensory responses from the endothelial
cells across whose surfaces the nanorobots touch~\cite{freitas03}.
If the aggregates cover a long section of the vessel wall, they
could produce local edemas since narrowing of the vessels by the
presence of the nanorobots increases local pressure gradients and
fluid velocities. While we focus on a single aggregate in one
microscopic vessel, additional issues arise if a large population of
circulating robots form many aggregates. In that case, the multiple
aggregates will increase hydrodynamic resistance throughout the
fluidic circuit. Thus the robots could make the heart work slightly
harder to pump fluid against the slightly higher load. Moreover, if
robot aggregates detach from the wall without complete
disaggregation, these smaller aggregates moving in the blood may be
large enough to block a small vessel.

The scenarios examined in this paper can suggest suitable controls
to distribute power when robots aggregate. Moreover, power control
decisions interact with the choices made for the aggregation
process. For example, if the task requires a certain amount of total
power for the aggregate (e.g., as a computation hub) then the
aggregation self-assembly protocol would depend on how much oxygen
is available, e.g., to make a larger aggregate in vessels with less
available oxygen, or recruit more passing robots when the task needs
more power. An example of this latter case could be if aggregates
are used as computation hubs to validate responses to rare events:
when local sensor readings indicate the possibility of such an
event, the aggregate could temporarily recruit additional robots to
increase power and computational capability for evaluating whether
those readings warrant initiating treatment.

Another approach to designing controls for teams of robots is the
formalism of partially observable Markov
processes~\cite{cassandra94,seuken08}. This formalism allows for
arbitrarily complex computations among the robots to update their
beliefs about their environment and other robots. Unfortunately this
generality leads to intractable computations for determining optimal
control processes. For the situations we studied, the power
constraints on capillary wall-resident microscopic robots operating
with oxygen available \textit{in vivo} means the local rules must be
simple. Including this constraint in the formalism could allow it to
identify feasible control choices for large aggregates of
microscopic robots in these situations.

The power constraints from our model could provide useful parameters
for less detailed models of the behavior of large numbers of robots
in the circulation in the context of the scenarios examined in this
paper. In particular, power limits the computation, communication
and locomotion capabilities of the robots. These constraints could
be incorporated in simplified models, such as cellular automata
approaches to robot behavior. These automata are a set of simple
machines, typically arranged on a regular lattice. Each machine is
capable of communicating with its neighbors on the lattice and
updates its internal state based on a simple rule. For example, a
two-dimensional scenario shows how robots could assemble
structures~\cite{arbuckle04} using local rules. Such models can help
understand structures formed at various scales through simple local
rules and some random motions~\cite{whitesides02,griffith05}. A
related analysis technique considers swarms~\cite{bonabeau99}, i.e.,
groups of many simple machines or biological organisms such as ants.
In these systems, individuals use simple rules to determine their
behavior from information about a limited portion of their
environment and neighboring individuals. Typically, individuals in
swarms are not constrained to have a fixed set of neighbors but
instead continually change their neighbors as they move. Swarm
models are well-suited to microscopic robots with their limited
physical and computational capabilities and large numbers. Most
swarm studies focus on macroscopic robots or behaviors in abstract
spaces~\cite{gazi04}. In spite of the simplified physics, these
studies show how local interactions among robots lead to various
collective behaviors~\cite{vicsek95} and provide broad design
guidelines. A step toward more realistic, though still tractable,
models of large aggregates could incorporate the power constraints
from the model presented in this paper.

In addition to evaluating performance of hypothetical medical
nanorobots, theoretical studies identifying tradeoffs between
control complexity and hardware capabilities can aid future
fabrication. One example is the design complexity of the robot's
fuel acquisition and utilization systems. For steady-state operation
on vessel walls, we found limited benefit of pumps over free
diffusion when numerous onboard power generators can be employed. In
such cases, our results indicate that a design without pumps does
not sacrifice much performance. More generally, control can
compensate for limited hardware (e.g., sensor errors or power
limitations), providing design freedom to simplify the hardware
through additional control programs. Thus the studies could help
determine minimum hardware performance capabilities needed to
provide robust systems-level behavior.

A key challenge for robot design studies based on approximate models
is validating the results. In our case, the most significant
approximations are the treatment of cells as an averaged component
in the fluid and the lumped-model kinetics for oxygen unloading.
With increased computational power, numerical solution of more
accurate models could test the validity of these approximations. As
technology advances to constructing early versions of microscopic
robots, experimental evaluations will supplement theoretical
studies.
One such experiment is operating the robots in manufactured
microfluidic channels~\cite{squires05}. This would test the robots'
ability to aggregate at chemically defined locations and generate
power reliably from known chemical concentrations in the fluid.
After such \textit{in vitro} experiments, early \textit{in vivo}
tests could involve robots acting as passive sensors in the
circulatory system and aggregating at chemically distinctive
locations. Such nanorobots will be useful not only as diagnostic
tools and sophisticated extensions to drug delivery
capabilities~\cite{allen04}, but also as an aid to develop robot
designs and control methods for more active tasks.

\small
\section*{Acknowledgements}
RAF acknowledges private grant support for this work from the Life
Extension Foundation and the Institute for Molecular Manufacturing.

\end{document}